\documentclass[10pt,twocolumn,letterpaper]{article}
\makeatletter
\def\@fnsymbol#1{\ensuremath{\ifcase#1\or \dagger\or \ddagger\or
   \mathsection\or \mathparagraph\or \|\or **\or \dagger\dagger
   \or \ddagger\ddagger \else\@ctrerr\fi}}
    \makeatother
\usepackage{iccv}              % To produce the CAMERA-READY version
\usepackage{times}
\usepackage[accsupp]{axessibility}
\usepackage{epsfig}
\usepackage{graphicx}
\usepackage{amsmath}
\usepackage{amssymb}
\usepackage{subcaption}
\usepackage{booktabs}
\usepackage{multirow}
\usepackage{amsthm}
\usepackage{enumitem}
\usepackage{colortbl}
\usepackage{bm}
\usepackage{framed}

\usepackage{capt-of}
\usepackage[dvipsnames]{xcolor}
\usepackage{bbding}
\newcommand{\modelname}{{{METEOR}}\xspace}
\definecolor{iccvblue}{rgb}{0.21,0.49,0.74}
\definecolor{custom_blue}{RGB}{235,244,253}
\usepackage[pagebackref,breaklinks,colorlinks,allcolors=iccvblue]{hyperref}
\def\paperID{686} % *** Enter the Paper ID here
\def\confName{ICCV}
\def\confYear{2025}
\begin{document}
% 备选
% \title{ME-CTP: Multi-Encoder Collaborative Token Pruning for Efficient Vision Language Models}
\title{\includegraphics[width=0.055\textwidth]{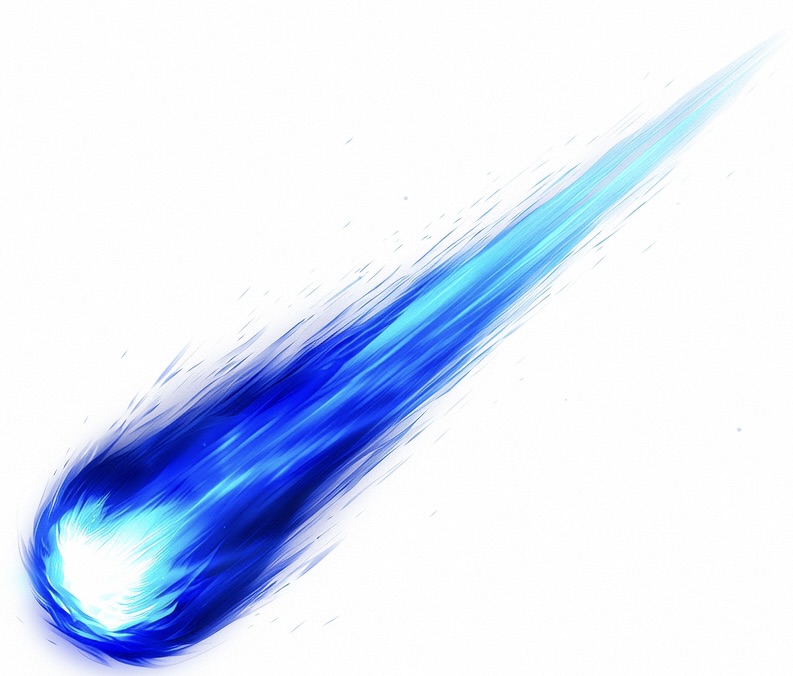}METEOR: Multi-Encoder Collaborative Token Pruning for Efficient \\ Vision Language Models}

\author{
Yuchen Liu$^{1}$, Yaoming Wang$^{3\dagger}$, Bowen Shi$^{1}$, Xiaopeng Zhang$^{2\dagger}$, Wenrui Dai$^{1}$\thanks{Corresponding author: Yaoming Wang, Xiaopeng Zhang, Wenrui Dai. This work was done when Yuchen Liu interned at Huawei Inc.}, Chenglin Li$^{1}$, \\ Hongkai Xiong$^1$ and Qi Tian$^2$\\
$^1$Shanghai Jiao Tong University, China $^2$Huawei Inc., China $^3$Meituan Inc., China \\
{\tt\small\{liuyuchen6666, sjtu\_shibowen, daiwenrui, lcl1985, xionghongkai\}@sjtu.edu.cn} \\{\tt\small wangyaoming03@meituan.com}, {\tt\small zxphistory@gmail.com}, {\tt\small tian.qi1@huawei.com}}

\maketitle
\begin{abstract}
Vision encoders serve as the cornerstone of multimodal understanding. Single-encoder architectures like CLIP exhibit inherent constraints in generalizing across diverse multimodal tasks, while recent multi-encoder fusion methods introduce prohibitive computational overhead to achieve superior performance using complementary visual representations from multiple vision encoders. To address this, we propose a progressive pruning framework, namely \textbf{M}ulti-\textbf{E}ncoder Collabora\textbf{T}iv\textbf{E} t\textbf{O}ken p\textbf{R}uning (\textbf{METEOR}), that eliminates redundant visual tokens across the encoding, fusion, and decoding stages for multi-encoder MLLMs. For multi-vision encoding, we discard redundant tokens within each encoder via a rank guided collaborative token assignment strategy. Subsequently, for multi-vision fusion, we combine the visual features from different encoders while reducing cross-encoder redundancy with cooperative pruning. Finally, we propose an adaptive token pruning method in the LLM decoding stage to further discard irrelevant tokens based on the text prompts with dynamically adjusting pruning ratios for specific task demands. To our best knowledge, this is the first successful attempt that achieves an efficient multi-encoder based vision language model with multi-stage pruning strategies. Extensive experiments on 11 benchmarks demonstrate the effectiveness of our proposed approach. Compared with EAGLE, a typical multi-encoder MLLMs, \textbf{METEOR} reduces 76\% visual tokens with only 0.3\% performance drop in average. The code is available at \url{https://github.com/YuchenLiu98/METEOR}.
\end{abstract}

\begin{figure}
    \centering
    \includegraphics[width=0.99\linewidth]{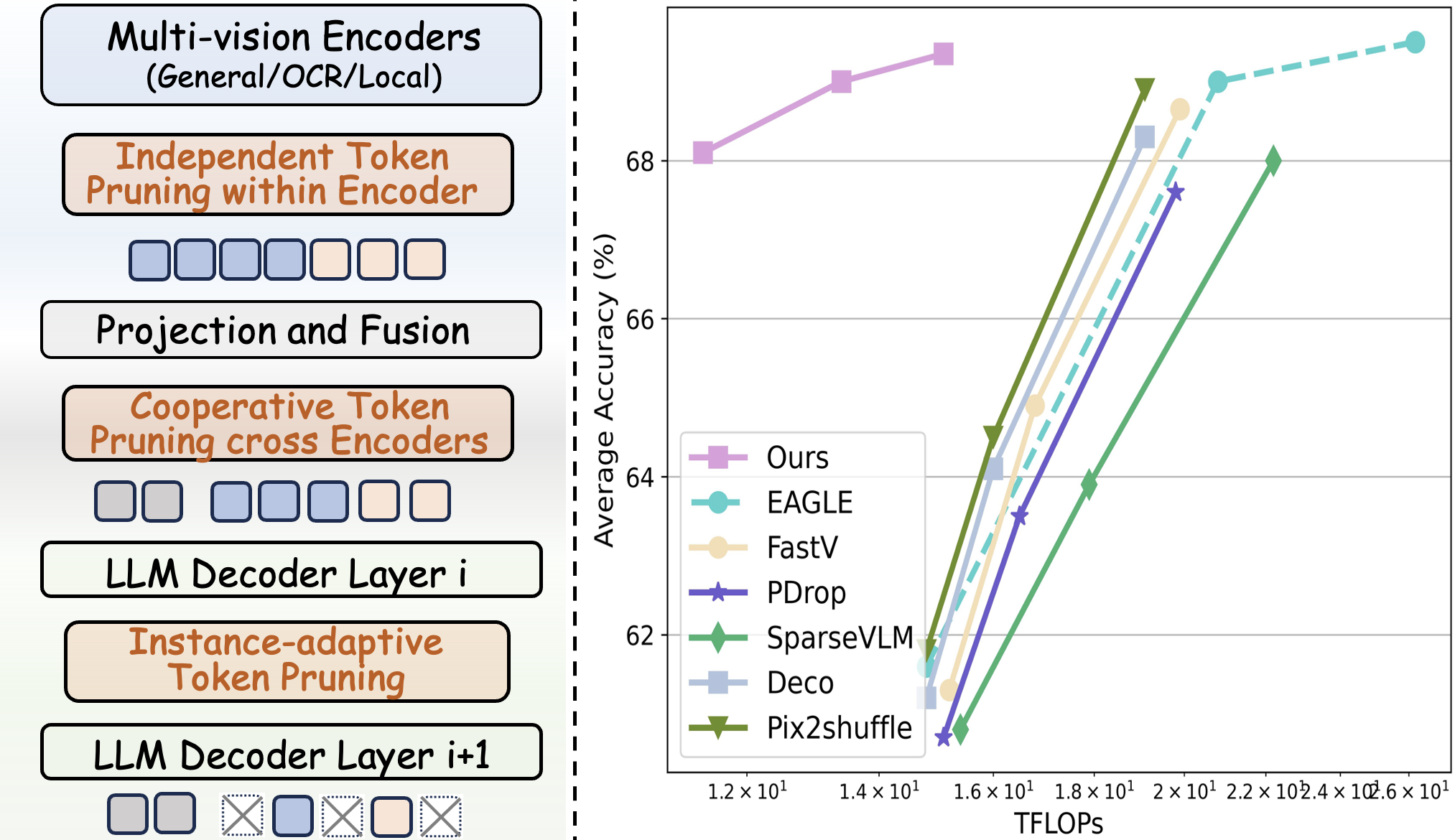}
    \vspace{-0.15cm}
    \caption{Overview of \modelname, which collaboratively prune redundant visual tokens across the encoding, fusion and decoding process of Muti-encoder MLLMs. Comparison with existing efficient methods based on EAGLE-X4~\cite{shi2024eagleexploringdesignspace} regarding accuracy and flops. Other two results of EAGLE are with one and two encoders.}
    \vspace{-0.4cm}
    \label{fig:accuracyvsflops}
\end{figure}
\vspace{-0.6cm}
\section{Introduction}
\vspace{-0.1cm}
Recent advancements in large language models (LLMs)~\cite{bai2023qwentechnicalreport,brown2020language,dai2019transformer,touvron2023llama} have significantly propelled the development of multi-modal LLMs (MLLMs), enabling promising applications that bring artificial intelligence into daily life~\cite{Qwen-VL,chen2024internvlscalingvisionfoundation,li2024minigeminiminingpotentialmultimodality,liu2023visual,zhu2023minigpt}. 
% Recently, benefiting from impressive advancements in large language models (LLMs)~\cite{bai2023qwentechnicalreport,brown2020language,dai2019transformer,touvron2023llama}, multi-modal LLMs (MLLMs) have demonstrated tremendous progress and enabled promising applications in bringing artificial intelligence into daily life~\cite{Qwen-VL,chen2024internvlscalingvisionfoundation,li2024minigeminiminingpotentialmultimodality,liu2023visual,zhu2023minigpt}. 
Typical MLLMs transform images into sequences of visual tokens, and then concatenate them with text tokens for LLMs. Subsequent works further enhance performance through high-resolution inputs~\cite{li2024monkeyimageresolutiontext,xu2024llavauhdlmmperceivingaspect} and innovative transformer architectures~\cite{Qwen-VL,tong2024cambrian1fullyopenvisioncentric}.
However, due to the inherently sparse nature of visual information compared to textual information, directly concatenating visual tokens causes significant computational inefficiency.
To mitigate this inefficiency, prompt-agnostic methods reduce the number of visual tokens before feeding into LLMs for single-encoder MLLMs like LLaVA~\cite{liu2023visual}. 
Visual tokens are compressed by visual attention mechanisms~\cite{jiang2024fopru,zhang2024cls,shang2024llava}, local pooling~\cite{li2024minigeminiminingpotentialmultimodality,yao2024deco,cai2024matryoshka,ding2023prune}, and resamplers~\cite{xu2024llavauhdlmmperceivingaspect}. However, these methods overlook the context provided by text prompts, resulting in redundant visual tokens irrelevant to user instructions. As a remedy, prompt-aware methods~\cite{yinunraveling,xing2024pyramiddrop,ye2024fit,hu2024illava} such as FastV~\cite{chen2025image}, Pdrop~\cite{xing2024pyramiddrop} and SparseVLM~\cite{zhang2024sparsevlm} employ text prompts to guide the visual token pruning in LLMs.
% To combine visual signals with textual semantics, MLLMs typically represent images as sequences of visual tokens, which are concatenated with text tokens and fed into LLMs. 

\begin{figure*}[!t]
    \centering
\includegraphics[width=1.0\linewidth]{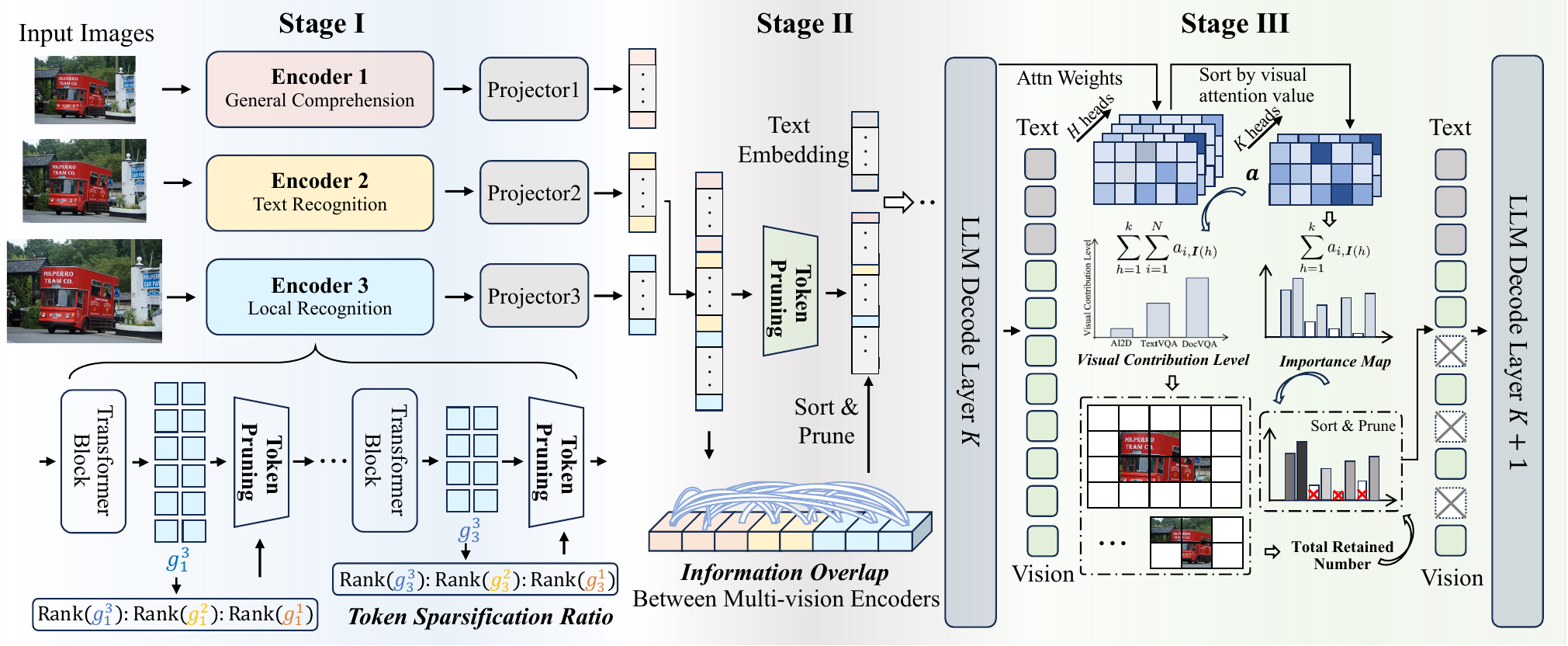}
    \vspace{-5.7mm}
    \caption{Overall architecture of \modelname. Redundant visual tokens are pruned within each encoder during encoding in stage 1. The rank of output feature map is utilized to assess the information richness for allocating the token sparsity ratio of each encoder. 
    Besides, based on the independent projector for adapting tokens to the shared semantic space, we prune mutually redundant visual tokens with information overlap across multi-encoders. Finally, we employ the vision-text attention value of the top-$k$ most relevant attn heads to precisely identify redundant tokens, and dynamically adjust the retained token number for varying task demands based on visual contribution level.}
    \label{fig:overall}
    \vspace{-1mm}
\end{figure*}

On the other hand, widely adopted single-encoder architectures like CLIP~\cite{radford2021learning} suffer from hallucinations~\cite{li2024monkeyimageresolutiontext,liu2024llavanext} and are degraded on fine-grained tasks such as grounding and OCR~\cite{jiang2024clipdinovisualencoders,wei2024vary}. 
To address these limitations, recent works~\cite{kar2024brave,zong2024mova,shi2024eagleexploringdesignspace,tong2024cambrian1fullyopenvisioncentric,fan2024mousipolyvisualexpertvisionlanguagemodels,liu2024prismervisionlanguagemodelmultitask} 
explore integrating multiple state-of-the-art vision encoders to improve visual perception for robust performance across diverse domains.
% have attempted to integrate multiple state-of-the-art vision encoders to enhance the visual perception for better all-around performance across diverse domains.
Despite promising performance, multiple vision encoders with high-resolution inputs introduce prohibitive computational overhead. For instance, in Mini-Gemini~\cite{li2024minigeminiminingpotentialmultimodality}, a 672$\times$672 image processed by dual vision encoders generates 2880 visual tokens, leading to quadratic scaling of computational complexity in self-attention mechanisms. 

% Since visual information is more sparse than textual information, there are much redundancy in visual tokens.
% Visual information is inherently more sparse than textual information, leading to significant redundancy in visual tokens. 
% To mitigate this inefficiency, prompt-agnostic methods propose to reduce the number of visual tokens before feeding into LLMs for single-encoder MLLMs such as LLaVA~\cite{liu2023visual}. 
% Techniques such as visual attention mechanisms~\cite{jiang2024fopru,zhang2024cls,shang2024llava}, local pooling~\cite{li2024minigeminiminingpotentialmultimodality,yao2024deco,cai2024matryoshka}, and resamplers~\cite{xu2024llavauhdlmmperceivingaspect} have been employed to compress visual tokens.
% However, these methods often overlook the context provided by text prompts, resulting in the retention of visual tokens that are irrelevant to user instructions. As a remedy, prompt-aware methods~\cite{yinunraveling,xing2024pyramiddrop,ye2024fit,hu2024illava}, including FastV~\cite{chen2025image}, Pdrop~\cite{xing2024pyramiddrop} and SparseVLM~\cite{zhang2024sparsevlm}, propose to leverage text prompts to guide the vision token pruning in LLMs.

To tackle this problem, one intuitive approach is to apply single-enoder token pruning strategies in multi-encoder MLLMs.
However, such an approach falls short in achieving collaborative pruning across multiple vision encoders. Specifically, it fails to address critical issues such as the appropriate allocation of token pruning ratios to vision encoders with varying levels of semantic richness and effective reduction of redundant tokens that overlap across multiple vision encoders.
% Despite promising performance, existing methods cannot be directly applied to accelerate multi-vision experts MLLMs. 
% Since the vision encoding takes up much time as the resolution increases shown in Fig.~\ref{fig:sec32} (a), how to accelerate the entire process across the encoding, fusion and decoding is important. Besides, how to appropriately allocate token sparsification ratio to different vision experts of varying semantic richness and mitigate the mutual redundant tokens across multi-vision experts are crucial for better accuracy-efficiency trade off. 
It is also crucial to accelerate the whole process under the complexity of vision encoding increasing with image resolution in \figurename~\ref{fig:sec32}(a).
%Besides, since the vision encoding takes much time with incrased resolution in \figurename~\ref{fig:sec32}(a), how to accelerate the whole process is crucial.
Moreover, existing pruning methods perform poorly on some specific tasks like OCR recognition due to their inability to dynamically adjust pruning ratio for varying task demands, since OCR tasks require retaining more tokens than general tasks. 
% For instance, they fail to be instance-adaptive with dynamically adjusted pruning ratio for OCR demands, i.e., fewer retained tokens for general and more for OCR tasks.
% For instance, they perform poorly on OCR tasks, since they adopt a fixed pruning ratio for both general and OCR tasks
% fail to be instance-adaptive with dynamically adjusted pruning ratio for OCR demands, i.e., fewer retained tokens for general and more for OCR tasks.
% Moreover, existing token pruning methods perform poorly on OCR tasks, since they fail to be instance-adaptive with dynamically adjusted pruning ratio for varying task demands, e.g., fewer retained tokens for general and more for OCR tasks.

% To this end, this paper proposes \textbf{M}ulti-\textbf{e}xpert adap\textbf{t}iv\textbf{e} t\textbf{o}ken spa\textbf{r}sification (\textbf{Meteor}) instantiated by a multi-stage token sparsity strategy
In this paper, we propose a novel multi-modal token pruning framework, namely \textbf{M}ulti-\textbf{E}ncoder Collabora\textbf{T}iv\textbf{E} t\textbf{O}ken p\textbf{R}uning (\textbf{METEOR}) to progressively eliminate redundant visual tokens across the whole process of multi-encoder MLLMs, including encoding, fusion and decoding. Specifically, for multi-vision encoding, we identify reliable criterion that measures similarity to the average token in shallow layers and class attention in deep layers for progressive pruning of redundant visual tokens. To allocate the appropriate token sparsity ratio for each encoder, we leverage the rank of feature maps produced by different encoders as the mathematically grounded measure of information richness. 
% The ratio derived from this measure is then used for allocation.
Furthermore, we find that the average rank of feature maps in a layer of the encoder is robust to input images with small variance, and provide tractable offline rank computation on a small batch with negligible computation overhead. 

For multi-vision encoder fusion, we employ a more flexible fusion strategy to maintain its own dedicated projector for each vision encoder, enabling independent adapting of visual tokens before fusion. Based on the adapted visual embeddings, we propose a cooperative pruning strategy to reduce mutually redundant visual tokens with information overlap across multi-vision encoders. 
Finally, for LLM decoding, we employ visual-text attention to reduce redundant visual tokens irrelevant to the text prompt. Different from existing methods averaging all the attention heads, we select top-$k$ significant heads to mitigate hallucination. Leveraging the correlation between the visual attention value and instance complexity across datasets, we dynamically adjust the token pruning ratio for specific tasks like OCR. 

To our best knowledge, METEOR is the first token pruning framework tailored for multi-encoder MLLMs. The contributions are summarized as:
\begin{itemize}
    % \item We propose a multi-stage framework to prune redundant vision tokens for multi-vision encoder MLLMs across encoding, fusion and decoding procedures. 
    \item We propose a multi-encoder collaborative token pruning framework to  progressively eliminate redundant visual tokens across the encoding, fusion, and decoding stages.
    % \item We propose to leverage the rank of feature maps output by different encoders to measure the information richness with mathematical proof, whose ratio guides the allocation of retained token number for various experts.
    \item We propose to utilize the rank of feature maps as a mathematically grounded measure of information richness to collaboratively allocate sparsity ratios for each encoder, and prune mutually redundant tokens with information overlap across multi-encoders during fusion.
    \item We develop an instance-adaptive pruning strategy that selects most relevant attention heads for accurate visual-text attention to identify redundant tokens, and further dynamically adjusts pruning ratios for varying task demands.   
\end{itemize}

Extensive experiments on multimodal benchmarks demonstrate the effectiveness of \modelname. Compared with state-of-the-art multi-encoder MLLMs EAGLE, \modelname reduces 76\% visual tokens while saving 49\% TFLOPS and accelerating 46\% FPS with only 0.3\% performance drop. Besides, our method consistently outperforms prior token pruning methods by 4.3\text{-}5.7\% on average, especially 8.8\text{-}12.3\% on OCR tasks, exhibiting the advantage of our instance-adaptive token pruning. Moreover, with the same vision encoders, our method significantly outperforms Cambrain-1~\cite{tong2024cambrian1fullyopenvisioncentric} by 3.3\% on OCR tasks with 44\% fewer visual tokens, demonstrating the broad applicability of our strategy across various multi-encoder MLLMs.

\label{sec:intro}
% \vspace{-0.15cm}
\section{Related Work}
% \vspace{-0.07cm}
\textbf{Multi-modal Large Language Models (MLLMs).} In recent years, MLLMs~\cite{Qwen-VL,chen2024internvlscalingvisionfoundation,li2024minigeminiminingpotentialmultimodality,tong2024cambrian1fullyopenvisioncentric,liu2023visual} have achieved promising progress. 
% With the impressive advancement of large language models (LLMs)~\cite{bai2023qwentechnicalreport,brown2020language,dai2019transformer,touvron2023llama}, growing attention has been drawn in integrating the vision signals into LLMs, which leads to the development of powerful multi-modal LLMs (MLLMs)~\cite{Qwen-VL,chen2024internvlscalingvisionfoundation,li2024minigeminiminingpotentialmultimodality,liu2023visual}.
% Flamingo~\cite{alayrac2022flamingo} and BLIP-2~\cite{li2023blip} bridge the modality gap between pretrained LLMs and visual encoders with promising zero-shot visual-language performance.
% Flamingo~\cite{alayrac2022flamingo} and BLIP-2~\cite{li2023blip} bridge the modality gap between pretrained LLMs and visual encoders.
LLaVA~\cite{liu2023visual} and MiniGPT-4~\cite{zhu2023minigpt} conduct instruction tuning on high-quality datasets to enhance generation capabilities under complex instructions. For further improvement, later researches explore high-resolution inputs~\cite{li2024monkeyimageresolutiontext,liu2024llavanext,xu2024llavauhdlmmperceivingaspect}, model design~\cite{Qwen-VL,tong2024cambrian1fullyopenvisioncentric}, and scaling up model size and data~\cite{chen2024fargpt4vclosinggap,chen2024internvlscalingvisionfoundation,li2024minigeminiminingpotentialmultimodality}. However, processing high-resolution images inevitably results in an exponential increase in the length of visual tokens. Since the number of text tokens is much smaller, the overall computational cost grows quadratically with the number of visual tokens. Thus, how to reduce the redundant visual tokens while maintaining the performance is critical for improving efficiency.

\noindent \textbf{Vision Encoders Design for MLLMs.} Previous MLLMs usually adopt typical vision-language pretrained models CLIP~\cite{radford2021learning,sun2023eva}. Despite good abilities in general recognition, these models lack specific abilities such as reading text and localizing objects. Hence, a series of works~\cite{jiang2024clipdinovisualencoders,lee2024moaimixtureintelligencelarge,li2024minigeminiminingpotentialmultimodality,lin2023sphinx,liu2024prismervisionlanguagemodelmultitask,lu2024deepseekvlrealworldvisionlanguageunderstanding,tong2024eyeswideshutexploring,he2024incorporatingvisualexpertsresolve} have integrated vision models pre-trained on diverse vision tasks or vision-language tasks to extend the visual capabilities. For example, Mousi~\cite{fan2024mousipolyvisualexpertvisionlanguagemodels}, Brave~\cite{kar2024brave}, Cambrian-1~\cite{tong2024cambrian1fullyopenvisioncentric} and EAGLE~\cite{shi2024eagleexploringdesignspace} fuse visual tokens from different vision experts, e.g., OCR, detection and segmentation, by concatenating along the channel or token direction. MoVA~\cite{zong2024mova} proposes a routing network to select optimal vision model combinations based on the given instructions. 
% Valley2~\cite{wu2025valley2exploringmultimodalmodels} fuses visual tokens output by Qwen2-VL~\cite{wang2024qwen2vlenhancingvisionlanguagemodels} vision encoder and siglip along the token direction. 
Despite promising performance, the visual encoding time of multi-vision encoders increases dramatically, even several times LLM prefilling time. Thus, how to reduce the computation complexity of multi-vision encoders in MLLMs with an optimized accuracy-latency trade-off is an urgent issue.

\noindent \textbf{Visual Token Compression for MLLMs.}
% Visual tokens usually occupy the majority of the input sequence in MLLM. Compared with text, visual information has higher redundancy, making visual token compression in either visual encoder or LLM phase effecitive for acceleration. 
% Visual tokens usually occupy the majority of the input sequence in MLLM. Compared with text, visual information has higher redundancy, making visual token compression effecitive for accelerating MLLMs. 
Compared with text, visual tokens occupy the majority of the input sequence with high redundancy, making visual token compression effective for accelerating MLLMs. 
Prompt-agonistic methods~\cite{shang2024llava,wang2024cls,zhang2024cls,jiang2024fopru,arif2024hired,wang2025folder,huang2025ivtp} leverage attention mechanisms of the vision encoder to identify important visual tokens. 
% LLaVAPruMerge~\cite{shang2024llava}, FasterVLM~\cite{wang2024cls}, VTC-CLS~\cite{zhang2024cls}, FoPru~\cite{jiang2024fopru}, HiRED~\cite{arif2024hired}, FOLDER~\cite{wang2025folder} and IVTP~\cite{huang2025ivtp} leverage attention mechanisms within the visual encoder to select important visual tokens. 
% Attention mechanisms within the visual encoder is leveraged to select important visual tokens~\cite{shang2024llava,wang2024cls,zhang2024cls,jiang2024fopru,arif2024hired,wang2025folder,huang2025ivtp}.
% DeepStack~\cite{meng2024deepstack} proposes to stack the visual tokens into the transformer layers of LLMs from bottom to top. 
DeCo~\cite{yao2024deco} and TokenPacker~\cite{li2024tokenpacker} reduce visual tokens with efficient visual projectors as pooling or local attention. Prompt-aware methods~\cite{yinunraveling,ye2024atp,ye2024fit,zhao2024accelerating,zhu2024focusllava,he2024zipvl,zhong2024aim} leverage the text semantics to guide visual token pruning in LLMs.
FastV~\cite{chen2025image} prunes unnecessary visual tokens based on the vision-text attention values. PDrop~\cite{xing2024pyramiddrop} and SparseVLM~\cite{zhang2024sparsevlm} progressively reduce the number of retained visual tokens as the LLM layers deepen. 
% ATP-LLaVA~\cite{ye2024atp}, FitPrune~\cite{ye2024fit}, G-search~\cite{zhao2024accelerating}, DyVTE~\cite{wu2024accelerating}, FocusLLaVA~\cite{zhu2024focusllava}, ZipVL~\cite{he2024zipvl} and SparseVLM~\cite{zhang2024sparsevlm} instance-wise adaptively prune visual tokens in LLMs based on their attention values with text tokens. 
% YOPO~\cite{zhang2024treatvisualtokenstext} and RedundancyLens~\cite{li2025redundancylensrevealingexploitingvisual} propose to save the visual computation of self-attention and FFN.
The visual computation of self-attention and FFN is saved in~\cite{zhang2024treatvisualtokenstext,li2025redundancylensrevealingexploitingvisual}.
LLaVA-Mini~\cite{zhang2025llava} and VoCo-LLaMA~\cite{ye2024voco} explore extreme compression to one visual token. MQT~\cite{hu2024matryoshka} and M3~\cite{cai2024matryoshka} employ matryoshka representation learning to compress visual tokens. iLLaVA~\cite{hu2024illava}, Ficoco~\cite{han2024rethinking} and MustDrop~\cite{liu2024multi} propose to drop visual tokens both in vision encoding stage and LLM stage. Despite promising results, these methods cannot be directly applied to multi-encoder MLLMs, e.g., how to allocate token budgets to various vision encoders. 
% Besides, existing methods perform poorly on OCR tasks, since they fail to be instance-adaptive with dynamically adjusted token budget for varying task demands.
Besides, existing methods fail to adaptively adjust token budget for varying tasks, which perform poorly on OCR tasks.
% \vspace{-0.15cm}
\section{Methodology}
% \vspace{-0.07cm}
\subsection{Overview}
\figurename~\ref{fig:overall} depicts the proposed framework that progressively prunes redundant visual tokens in the whole process of multi-vision encoding, multi-vision fusion, and LLM decoding stages for multi-encoder MLLMs.

\begin{figure}[!t]
\centering
\renewcommand{\baselinestretch}{1.0}
\centering
\subfloat[Vision Encoding Latency.]{\includegraphics[width=0.48\linewidth,height=0.38\linewidth]{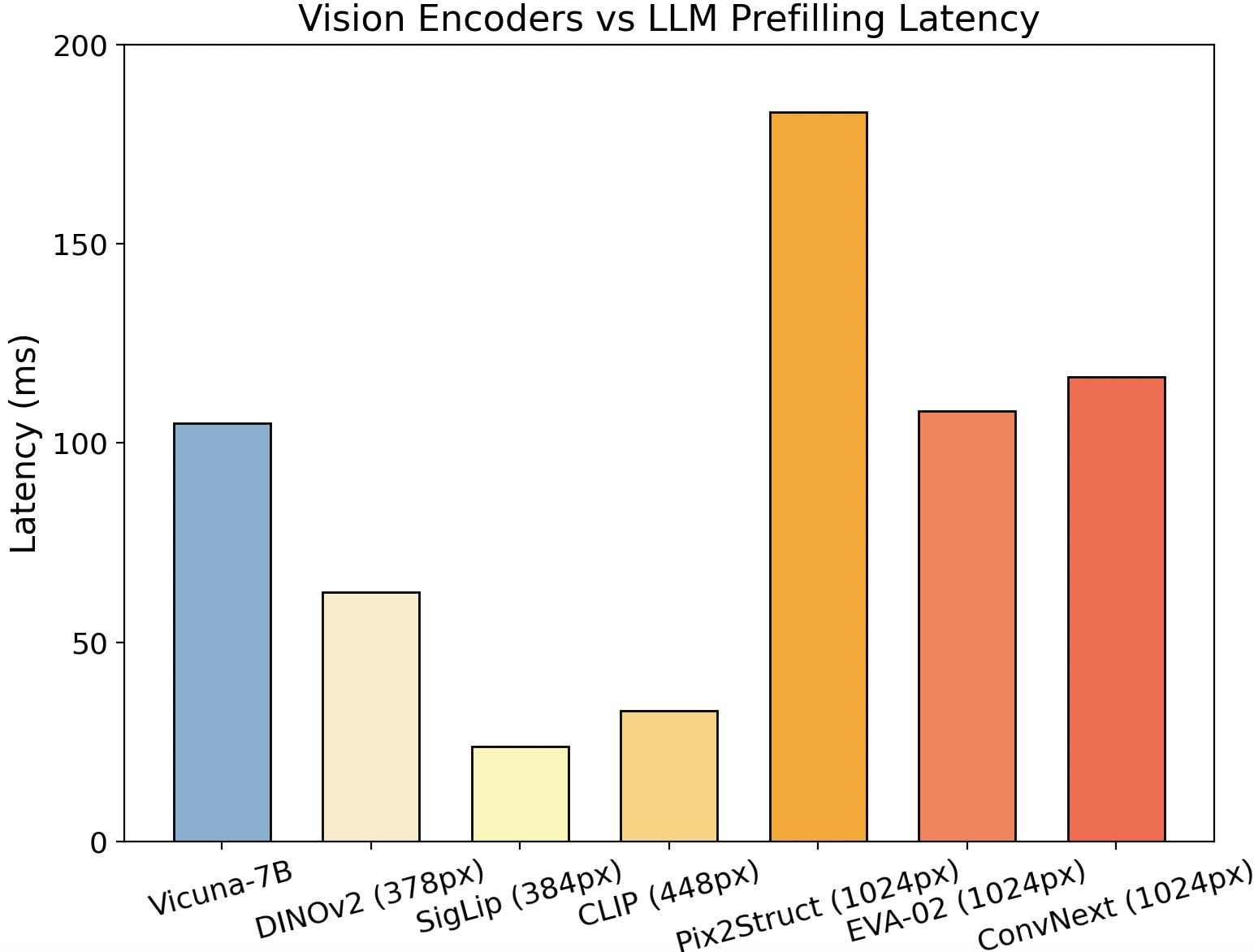}}
\hspace{2pt}
\subfloat[Entropy and Kendall Tau.]{\includegraphics[width=0.48\linewidth,height=0.38\linewidth]{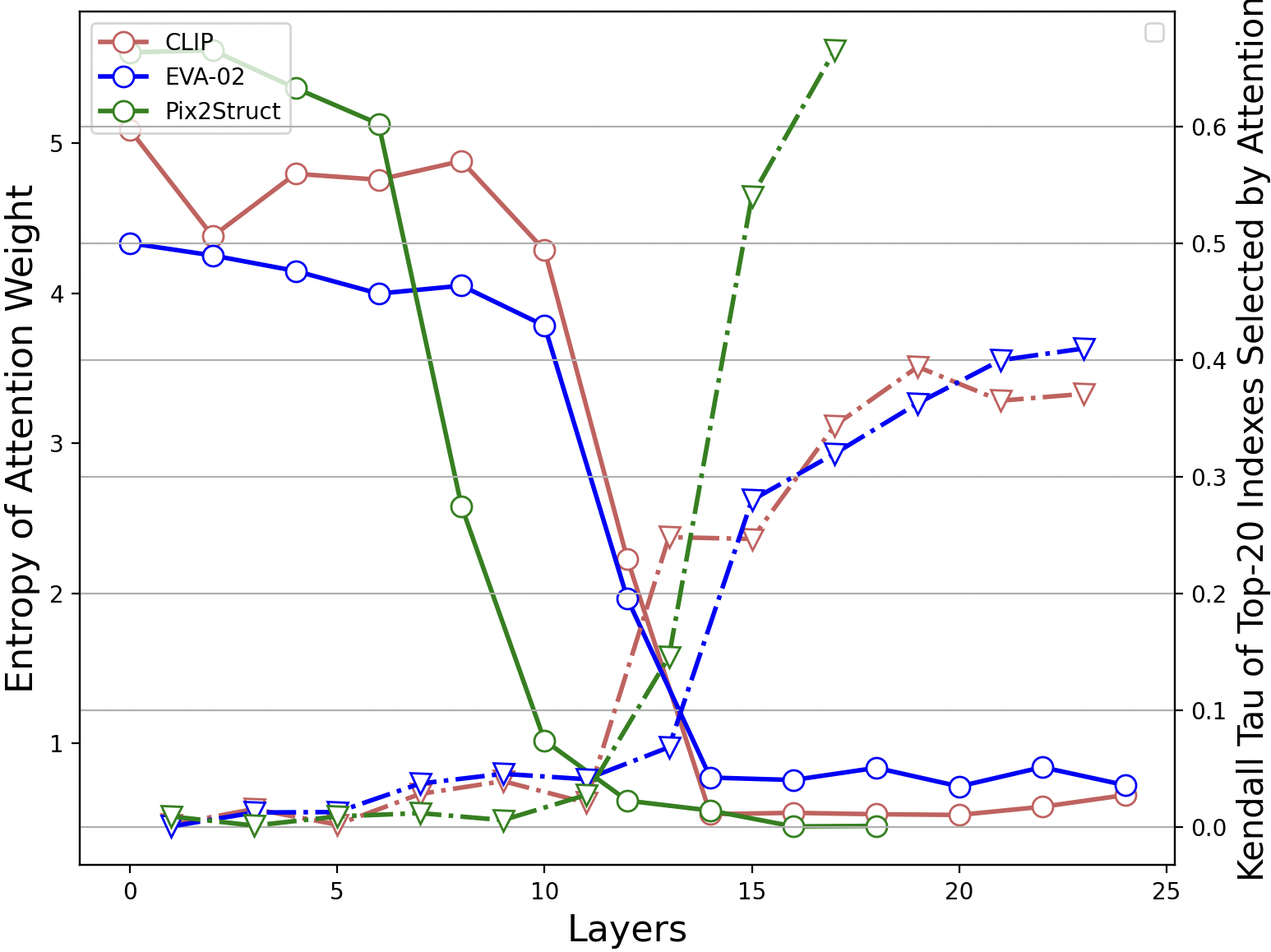}}
\vspace{-4pt}
\caption{(a) Vision encoding and LLM prefilling latency (fed with 576 tokens). (b) Entropy of attention value and kendall tau correlation of top-$k$ indices selected by attention values between adjacent layers.}\label{fig:sec32}
\vspace{-0.5cm}
\end{figure}

\noindent\textbf{Stage 1: Multi-vision Encoding.} \figurename~\ref{fig:sec32}(a) shows that the latency for vision encoding dramatically increases with the growth of image resolution and model complexity, and could even exacerbate for multi-vision encoders. We develop a reliable measure to identify and prune redundant tokens with a collaboratively allocated sparsity ratio.
%Since it is more costly with multi-vision encoders, we aim to accelerate the multi-vision encoding for better efficiency.

% Secondly, after independently dropping redundant tokens in each vision encoder, previous pre-adaptation fusion strategies~\cite{shi2024eagleexploringdesignspace,tong2024cambrian1fullyopenvisioncentric} fail to be applied due to the mismatching of retained tokens in each encoder. Besides, redundancy between visual tokens across multiple encoders should be further suppressed. Thus, how to achieve effective fusion while suppressing cross-encoder redundancy is important for multi-vision expert fusion.
\noindent \textbf{Stage 2: Multi-vision Fusion.} Existing methods~\cite{shi2024eagleexploringdesignspace,tong2024cambrian1fullyopenvisioncentric} adapt the independently pruned tokens from different encoders with one shared projector. They are not flexible and overlook mutually redundant tokens with information overlap across multi-encoders. We employ the dedicated projector to adapt tokens from each encoder and cooperatively suppress mutual redundancy across encoders during fusion.
%After independently dropping redundant tokens within each encoder, previous fusion methods~\cite{shi2024eagleexploringdesignspace,tong2024cambrian1fullyopenvisioncentric} are less flexible since they adopt one shared projector to adapt retained tokens from different encoders. Besides, there are mutually redundant tokens with information overlap across multi-encoders. Thus, we aim to achieve effective fusion and suppress the cross-encoder redundancy.

% In contrast, LEO employs an alternative fusion approach in which each vision encoder maintains its own dedicated projector module, allowing for independent processing of visual tokens before they are combined. We find this to be a more flexible and effective fusion
% strategy
\noindent \textbf{Stage 3: LLM Decoding.} Pruning using only vision information cannot remove redundancy related to specific text prompts, nor fulfill different token budget for instances of varying complexity, \emph{e.g.}, more tokens required for OCR than general tasks. We adaptively prune visual tokens with dynamically adjusted pruning ratio guided by text prompts.

% \subsection{METEOR}
\subsection{Stage 1: Independent Pruning within Encoders}
\label{sec:stage1}
% Given $l=1,\cdots L$ visual experts, where the $l$-th expert is stacked by $b=1,\cdots,B_l$ blocks, we propose to perform TOST in multiple blocks of each visual expert as discarding tokens all at once suffers serious information loss while progressively discarding at different blocks could largely alleviate it by exploiting the information propagation capability of the self-attention mechanism. 
% Then, for block $b$ of the $l$-th visual expert,
% we firstly figure out the indicator tokens $\bm{f}^{s_1,l,b}$, the cost matrix $\bm{C}^{s_1,l,b}$, and the sparse ratio $\gamma^{s_1,l,b}$.
Given $L$ vision encoders with the $l$-th encoder stacking $B_l$ blocks, we propose to progressively discard tokens in multiple blocks to sufficiently exploit the information propagation capability of self-attention mechanism and significantly alleviate the serious information loss by pruning all at once. For any $b$-th block in the $l$-th vision encoder, we figure out the redundant tokens and allocate a proper pruning ratio to each encoder with varying information richness.
\noindent\textbf{Redundant Token Identification.} Attention values are usually used to calculate the significance of each token~\cite{jiang2024fopru,shang2024llava,wang2024cls,zhang2024cls} but are not reliable in shallow layers.  
%Previous methods~\cite{jiang2024fopru,shang2024llava,wang2024cls,zhang2024cls} usually utilize the attention values to calculate the significance of each token. However, the attention values derived in shallow layers are not reliable. 
As shown in \figurename~\ref{fig:sec32}(a), the distribution of attention values is not sparse in shallow layers and exhibits high entropy. Besides, the top-$k$ indices selected by attention values vary significantly with low Kendall's tau correlation between adjacent layers. Since shallow layers usually contain low-level information, the average token corresponding to low-frequency components using discrete Fourier transform usually represents the background with high redundancy. Thus, we measure the significance with similarity to the average token as
% \begin{equation}
% \begin{aligned}
%  S_{\boldsymbol{v}} = &\mathcal{S}(\boldsymbol{v}, \sum^{N} \boldsymbol{v} / N) \in \mathbb{R}^{N}, \\
%  T(\boldsymbol{g_b^l}; k)=&\operatorname{gather}\left[\boldsymbol{v}, \operatorname{sort}(S_{\boldsymbol{v}})<k\right] \in \mathbb{R}^{k \times D}
% \end{aligned}
% \end{equation}
\vspace{-0.15cm}
\begin{equation}
 T(\boldsymbol{g_b^l}; k)\!=\!\operatorname{gather}[\boldsymbol{g_b^l}, \operatorname{sort}(-\mathcal{S}(\boldsymbol{g_b^l}, \sum^{n} \boldsymbol{g_b^l} / n))\!<\!k].
\end{equation}
Here, $\boldsymbol{g_b^l}\in\mathbb{R}^{N_b^{l}\times D}$ is the output tokens from the $b$-th block in the $l$-th vision encoder, $\operatorname{sort}$ means sorting descendingly, $\mathcal{S}(\cdot,\cdot)$ is the cosine similarity, and $T(\boldsymbol{g_b^l}; k)$ is the set of retained $k$ tokens. 
% \vspace{-0.2cm}
% \setlength{\FrameRule}{1pt} % 设置边框线的宽度
% \setlength{\FrameSep}{2pt} % 设置边框线与内容之间的间距
% \setlength{\leftmargin}{0pt} % 将左边框与左边距对齐
% \begin{framed}
% \noindent \textbf{Finding 1}: Since shallow blocks contain low-level redundant information, average token is more suitable for measuring the token significance compared to attention values.
% \end{framed}
% \vspace{-0.2cm}

% \vspace{0.03cm}
\noindent \emph{\textbf{Finding 1}: Average token is more suitable than attention values for measuring the token significance in shallow blocks containing low-level redundant information.}
%Since shallow blocks contain low-level redundant information, average token is more suitable for measuring the token significance compared to attention values.}
% \vspace{0.03cm}

Attention values are sparse and reliable for deep layers, as shown in \figurename~\ref{fig:sec32}(b). We measure the significance with attention values between the class token and visual tokens. 
\begin{equation}
T(\boldsymbol{g_b^l}; k)\!=\!\operatorname{gather}[\boldsymbol{g_b^l},\operatorname{sort}(\operatorname{softmax}({\mathbf{q}_{\mathrm{cls}}\!\cdot\!\mathbf{K}^{\mathbf{T}}}\!/\!{\sqrt{d_k}}))\!<\!k],
\end{equation}
where $\mathbf{q}^{\mathrm{cls}}$ is the query value of the cls token and $\mathbf{K}$ is the key value of $\boldsymbol{g_b^l}$.
%where $\mathbf{q}^{\mathrm{cls}}$ is the query value of the cls token and $\mathbf{K}$ is the key value of the feature tokens of the $b$-th block in the $l$-th encoder.
In practice, we equally split the model into three phases, where the first shallow phase employs the cosine similarity of features and last two phases employ the reliable attention value to identify redundant visual tokens.

\noindent\textbf{Sparsity Ratio for Multi-vision Encoders.}
Visual tokens from different vision encoders exhibit varying levels of semantic richness and redundancy, and differ in contributions to MLLM understanding. Here, we allocate varying sparsity ratios for each block to appropriately budget the number of retained tokens for semantic richness, \emph{i.e.}, fewer tokens for lower semantic richness.
%appropriately budgeting the number of retained tokens, i.e., the sparse ratio for each block of encoder is crucial, with fewer tokens allocated to encoders exhibiting low semantic richness, and vice versa. 
Inspired by~\cite{Lin_2020_CVPR}, we explore the rank of the feature map as a measure of token redundancy. The singular value decomposition (SVD) of $\boldsymbol{g_b^{l}}$ is
%For the output of the $b^{th}$ block in the $l^{th}$ vision encoder $\boldsymbol{g_b^{l}} \in \mathbb{R}^{N_b^{l} \times D}$, we conduct a Singular Value Decomposition (SVD) as
% \begin{align}
%     v_i^{j} &= \sum_{i=1}^{r}\sigma_iu_iw_i^\top \\ \nonumber
%     & = \sum_{i=1}^{r'}\sigma_iu_iw_i^\top + \sum_{i=r'+1}^{r}\sigma_iu_iw_i^\top
% \end{align}
\vspace{-0.18cm}
\begin{equation}
\bm{g_b^{l}} = \sum_{b=1}^{r}\sigma_b \alpha_b w_b^\top = \sum_{b=1}^{r'}\sigma_b \alpha_bw_b^\top + \sum_{b=r'+1}^{r}\sigma_b\alpha_bw_b^\top.
\vspace{-0.1cm}
\end{equation}
where $r$ is the rank of $\boldsymbol{g_b^{l}}$, and $\alpha_b$, $w_b$ are the top-$i$ left and right singular vectors. A feature map of rank $r$ is decomposed into a feature map $\sum_{b=1}^{r'}\sigma_b\alpha_bw_b^\top$ with lower rank $r'$ and additional information $\sum_{b=r'+1}^{r}\sigma_b\alpha_bw_b^{\top}$. This implies that rank is a reliable measure of information richness, since feature maps with higher rank contain more information.
%Hence, higher-rank feature maps actually contain more information than lower-rank ones, making rank a reliable measure of information richness. 

Since online computation of ranks is prohibitive, we use a small batch of input images to estimate the expectation of ranks in an offline manner. We extend the findings in~\cite{Lin_2020_CVPR} from CNNs to Vision Transformers and empirically reveal that the expectation of ranks is robust to the input images and the variance is negligible, as shown in \figurename~\ref{fig:sec2rank}(a). 
%However, computing the rank online incurs prohibitive complexity. Fortunately, we extend the findings in~\cite{Lin_2020_CVPR} from CNNs to Vision Transformers and empirically observe that the expectation of ranks is robust to the input images in \figurename~\ref{fig:sec2rank}(a).
% To illustrate, we plot the average of rank values w.r.t.
% different numbers of input batches in Fig. 2. It is obvious
% that the colors (which indicate the values) of the average
% ranks from a single filter are the same, which are irrespective
% to the image that CNNs receive. 
%To explain, although different images may have different ranks, the variance is negligible. Thus, we can employ a small batch of input images to estimate the expectation of the feature map rank offline.

\noindent \emph{\textbf{Finding 2}: The rank of feature map is a stable and reliable measure for the information richness of various encoders.}

% After obtaining the rank $r_i^j$, the budget of retained token number for the $i$-th layer in the $j$-th expert is computed as $k_i^j = r_i^j \cdot k_i/\sum_{j=1}^Cr_i^j$, where $k_i$ is the overall token number for the $i$-th layer.
The retained token number budget for the $b$-th layer of the $l$-th encoder is allocated proportionally to the rank $r_b^l$ as $k_b^l = k_b \cdot r_b^l/\sum_{l=1}^Cr_b^l$, where $k_b$ is the overall kept token number of the $b$-th block for all vision encoders.

% Considering that, we further quasi-, 
% $\sum_{i,j}P_{i,j}C_{i,j}=\sum_{j}p_j\sum_{i}C_{i,j}$.

% \subsection{Stage I: Token Redundancy in Multi-vision Encoding}
% \label{sec:stage1}
%  Since vision transformers are stacked by self-attention blocks, determining which layer to discard tokens is crucial. 
% \noindent\textbf{Degraded OT for Individual Vision Expert}

\subsection{Stage 2: Cooperative Pruning cross Encoders}
\textbf{Multi-vision Token Fusion.} 
% Existing multi-vision expert MLLMs~\cite{kar2024brave,luo2024feast,shi2024eagleexploringdesignspace} typically employ a pre-projection fusion strategy, combining visual tokens from various experts before the vision-text alignment with all visual encoders sharing the same projector. However, these early fusion strategies perform poor due to the mismatching of kept tokens after dropping separately in each encoder. In contrast, we employ a more flexible and effective fusion approach where each vision expert maintains its own dedicated projector, enabling independent adapting of visual tokens before fusion. Moreover, this strategy could first align visual embeddings from various experts, enabling to further reduce redundant tokens across multi-vision experts in the shared semantic space. 
Existing multi-encoder MLLMs \cite{kar2024brave,luo2024feast,shi2024eagleexploringdesignspace} employ a pre-projection fusion strategy to combine visual tokens before the vision-text alignment using a shared projector for all the vision encoders. In contrast, we adopt a more flexible post-projection fusion where each encoder maintains its own dedicated projector to independently adapt tokens for fusion. This strategy could align visual embeddings from various encoders to further reduce redundant tokens across multi-encoders in the shared semantic space. Specifically, we use a two-layer MLP $G$ as the projector, \emph{i.e.}, $\boldsymbol{z^j}\!=\!G(\boldsymbol{g^l})$, where $\boldsymbol{g^l}$ is the output of the $l$-th encoder. Visual tokens from multi-encoders are concatenated along the token channel as $\boldsymbol{z}=[ \boldsymbol{z^1},\!\cdots\!,\boldsymbol{z^L}]$.

\noindent\textbf{Mutual Redundancy across Multi-vision Encoders.}
% \textbf{Mutual Redundancy across Multi-vision Experts.} 
%Despite removing redundant tokens within each vision encoder in Section~\ref{sec:stage1}, mutual redundancy across various encoders still remains, requiring to be cooperatively pruned for better efficiency. One intuition is that tokens with mutual redundancy tend to have similar representations, which have analogous contributions in the following attention mechanism. 
Tokens with similar representations tend to have analogous contributions in the following attention mechanism and lead to mutual redundancy. We further achieve cooperative pruning to remove mutual redundancy across various vision encodes for better efficiency. For the $i$-th token $z^j_i$ in $\boldsymbol{z^j}$, the mutual redundancy is defined as:
% \begin{equation} \label{eq:simi}
%     \begin{aligned}
%     \mathcal{R}_i^j=&\sum_{l=1,l\neq j}^L\sum_{m=1}^{n_l}\mathcal{S}(z^j_i,z_m^l) \\
% T(\boldsymbol{z}; k)&=\operatorname{gather}[\boldsymbol{z}, \operatorname{sort}(-\boldsymbol{\mathcal{R}})<k]
% \end{aligned}
% \end{equation}
\vspace{-0.12cm}
\begin{equation} \label{eq:simi}
\mathcal{R}_i^j=\sum_{l=1,l\neq j}^L\sum_{m=1}^{n_l}\mathcal{S}(z^j_i,z_m^l).
\vspace{-0.08cm}
\end{equation}
We then keep the top-$k$ tokens with the lowest mutual redundancy and discard the rest as $T(\boldsymbol{z}; k)=\operatorname{gather}[\boldsymbol{z}, \operatorname{sort}(-\boldsymbol{\mathcal{R}})<k]$.
% \begin{equation}\label{eq:simi}
% \begin{aligned}
% \mathcal{R}_i^j=&\sum_{l=1,l\neq j}^n\sum_{m=1}^{k^l}\mathcal{S}(z^j_i,z_m^l) \\
% T(\boldsymbol{z^j} ; k)=&\operatorname{gather}\left[\boldsymbol{z^j}, \operatorname{sort}(\mathcal{R}_i^j)<k\right]  \\
% \end{aligned}
% \end{equation}
%where $\mathcal{S}(\cdot,\cdot)$ refers to cosine similarity. Then, the top-$k$ tokens with the lowest mutual redundancy are kept while the rest are discarded as $T(\boldsymbol{z}; k)=\operatorname{gather}[\boldsymbol{z}, \operatorname{sort}(-\boldsymbol{\mathcal{R}})<k]$.

\begin{figure}[!t]
\centering
\renewcommand{\baselinestretch}{1.0}
\centering
\subfloat[Average rank statistics.]{\includegraphics[width=0.5\linewidth,height=0.37\linewidth]{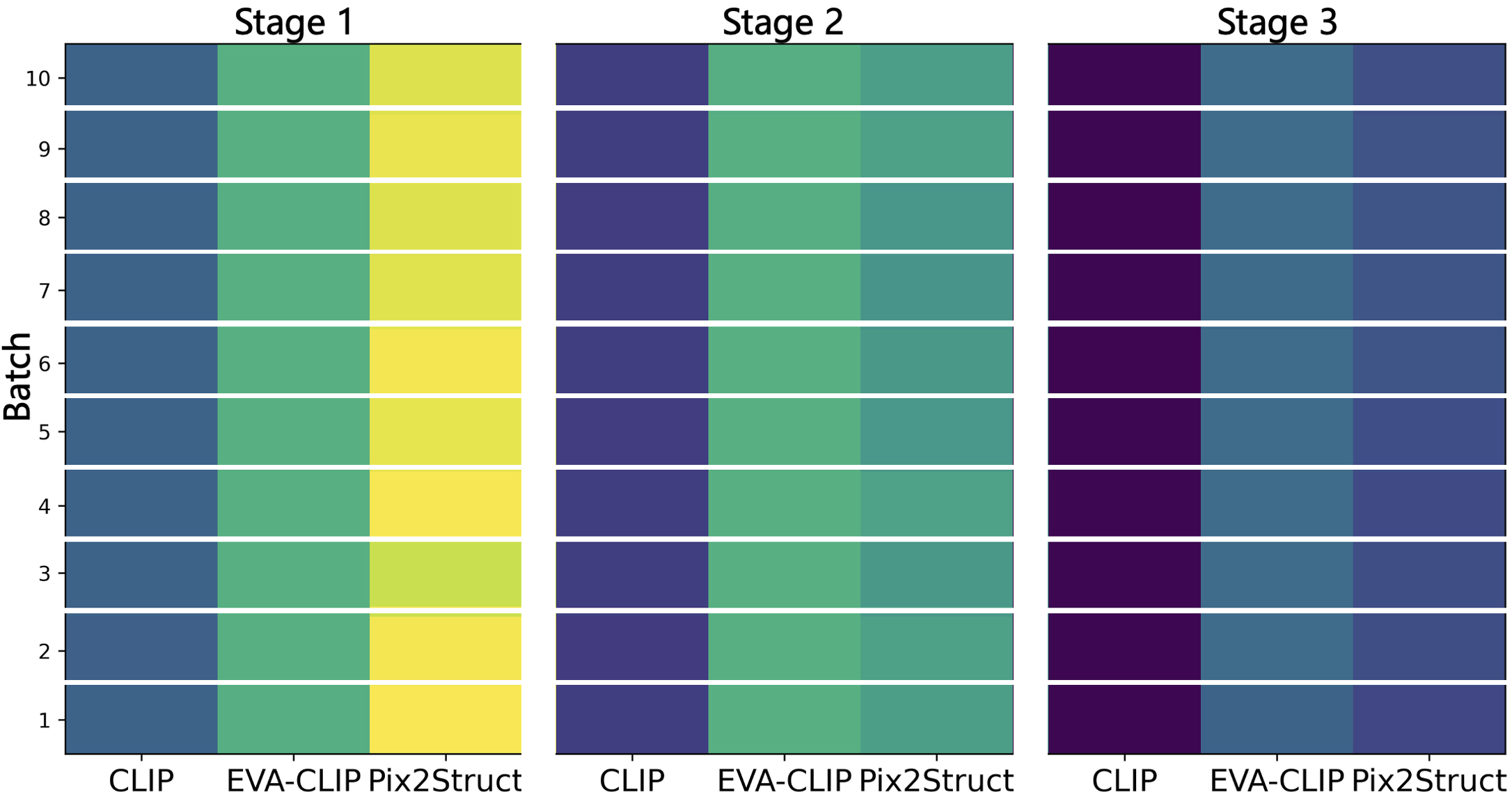}}
% \hspace{1pt}
\subfloat[Diversity of token sequences.]{\includegraphics[width=0.5\linewidth,height=0.37\linewidth]{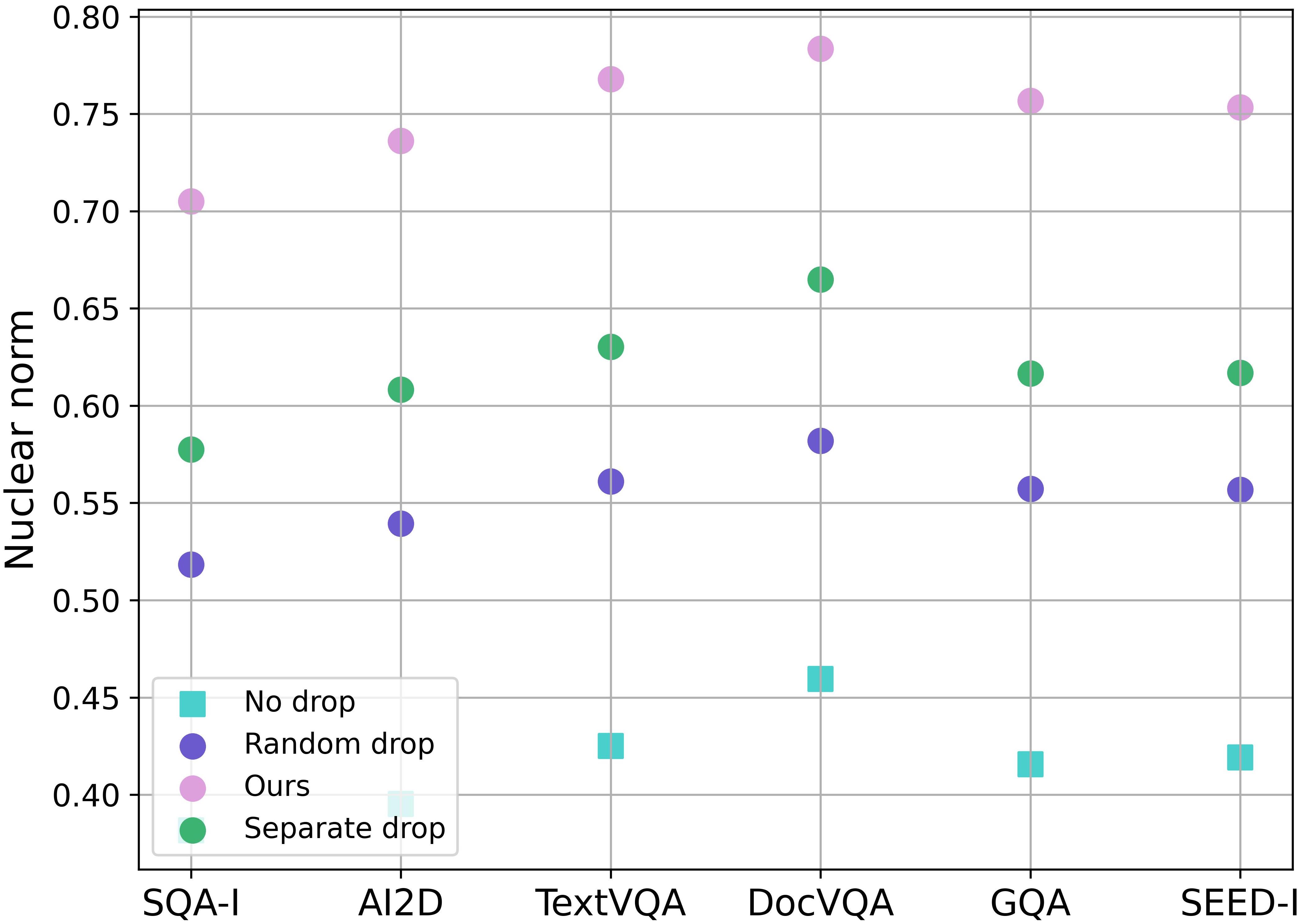}}
\vspace{-4pt}
\caption{(a) Rank statistics of feature maps across different phases of encoders, with color indicating rank values. As shown, the rank of each feature map hardly changes regardless of image batches. (b) Diversity of concatenated token sequences in stage 2.}\label{fig:sec2rank}
\vspace{-0.36cm}
\end{figure}

\noindent\textbf{Analysis on Diversity Enhancement.} We measure the feature diversity with the nuclear norm (more details in the supplementary material). \figurename~\ref{fig:sec2rank}(b) shows our cooperative pruning effectively enhances the diversity of token sequences compared to no or random discarding. Moreover, it outperforms separately dropping tokens in each encoder, highlighting the importance of reducing cross-encoder redundancy over within-encoder redundancy during fusion.
% \vspace{-0.23cm}
% \setlength{\FrameRule}{1pt} % 设置边框线的宽度
% \setlength{\FrameSep}{2pt} % 设置边框线与内容之间的间距
% \setlength{\leftmargin}{0pt} % 将左边框与左边距对齐
% \begin{framed}
% \noindent \textbf{Finding 2}: There are mutual redundancy across feature tokens of multi-vision experts, which requires to be further suppressed for better efficiency.
% \end{framed}
% \vspace{-0.32cm}

\noindent \emph{\textbf{Finding 3}: Reducing mutual redundancy across feature tokens of multi-vision encoders achieves better efficiency.}
%There are mutual redundancy across feature tokens of multi-vision encoders, which requires to be further suppressed for better efficiency.}

\subsection{Stage 3: Text-aware Instance-adaptive Pruning}
\textbf{Redundant Token Identification.} 
Redundant tokens can be further reduced considering specific text prompts beyond visual information and the sparsity ratio can be dynamically adjusted for varying complexity of input instances. Existing methods~\cite{chen2025image,hu2024illava,xing2024pyramiddrop,ye2024fit} suffer from degraded performance on fine-grained tasks using a pre-defined sparsity ratio. We identify redundant visual tokens with the attention value $a_{i}$ between the visual token $z_i$ and the last instruction token $t$. %where $a_{i,h}$ is the attention value of the $h$-th attention head. 
%After the removal of redundancy in previous stages based on the visual-only information, there are still redundant visual tokens not related to the specific text prompt. 
%Besides, due to varying complexity of input instances, a pre-defined pruning ratio in previous methods~\cite{chen2025image,hu2024illava,xing2024pyramiddrop,ye2024fit} leads to poor performance on fine-grained tasks, necessitating a dynamic adjustment of the retained token number. 
% Inspired by~\cite{chen2025image,xing2024pyramiddrop}, we adopt the attention value $a_{i}$ between visual token $z_i$ and last instruction token $t$ to identify redundant vision tokens as $a_{i}=\sum_{h=1}^{H}a_{i,h}$, where $a_{i,h}$ is the attention value of the $h$-th attention head. 
%In specific, we adopt the attention value $a_{i}$ between visual token $z_i$ and last instruction token $t$ to identify redundant visual tokens as $a_{i}=\sum_{h=1}^{H}a_{i,h}$, where $a_{i,h}$ is the attention value of the $h$-th attention head. 

\begin{figure}[!t]
\centering
\includegraphics[width=0.99\linewidth]{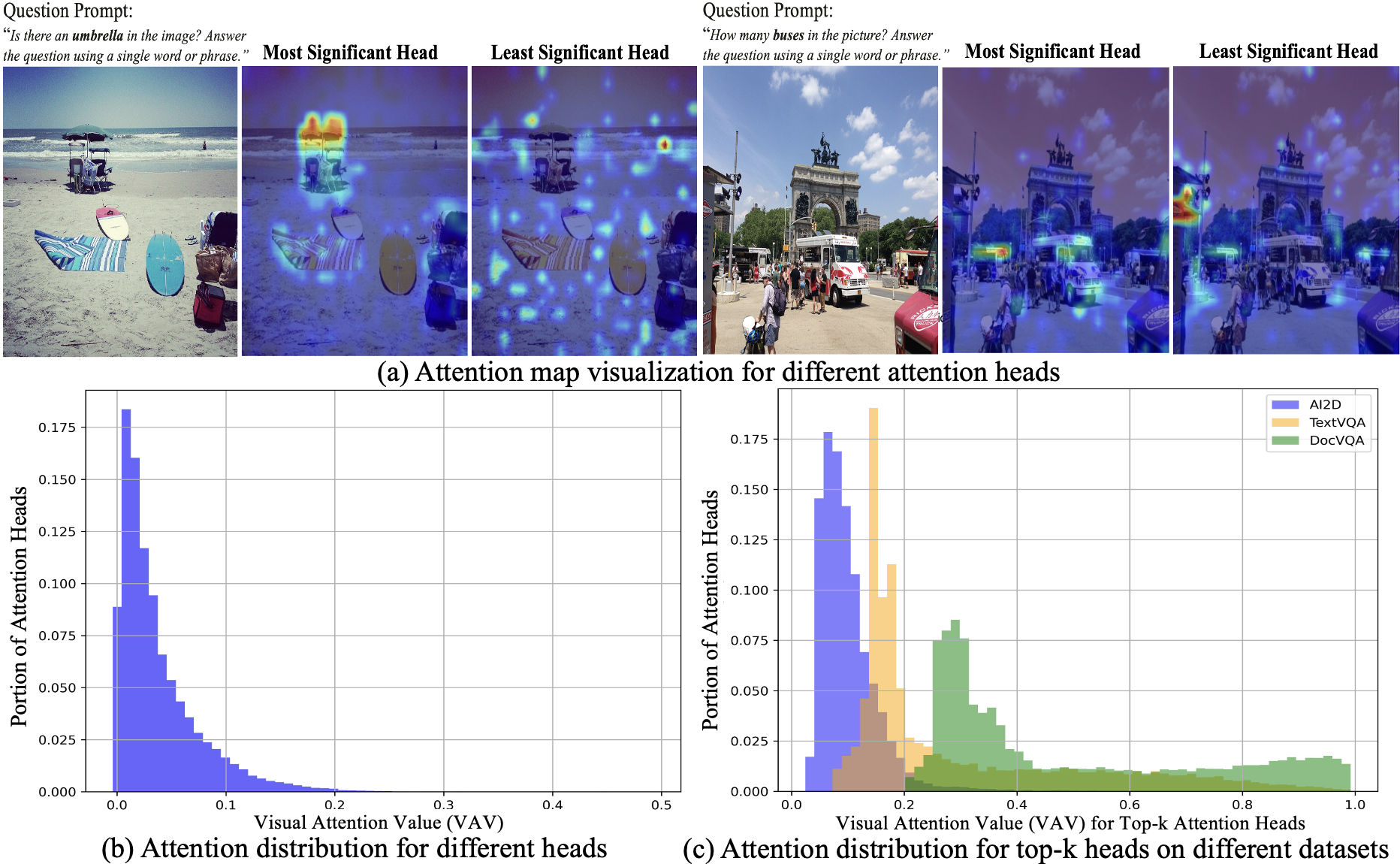}
\vspace{-2pt}
\caption{(a) Attention map visualization for different heads of the most significant value and the least value. (b) Visual attention value for different attention heads on AI2D. (c) Visual attention value for top-$k$ largest attention heads on different datasets.}\label{fig:figattention}
\vspace{-0.36cm}
\end{figure}

\noindent \emph{\textbf{Finding 4}: Not all attention heads are relevant for accurately identifying redundant visual tokens.}
% \vspace{-0.4cm}
% \setlength{\FrameRule}{1pt} % 设置边框线的宽度
% \setlength{\FrameSep}{2pt} % 设置边框线与内容之间的间距
% \setlength{\leftmargin}{0pt} % 将左边框与左边距对齐
% \begin{framed}
% \noindent \textbf{Finding 3}: Not all attention heads are relevant for accurately identifying redundant visual tokens.
% \end{framed}
% \vspace{-0.2cm}
% \textbf{Not All Attention Heads Are Equally Useful.} 

We adopt the Visual Attention Value (VAV) that is the magnitude of the attention value between the text token $t$ and all visual tokens to measure the quality of each attention head. For the $h$-th head, 
\vspace{-0.1cm}
\begin{equation}
    \operatorname{VAV}_h \triangleq \sum_{i=1}^{N} a_{i,h}
    \vspace{-0.1cm}
\end{equation}
where $a_{i,h}$ is the attention value of the $h$-th attention head. Existing methods~\cite{chen2025image,xing2024pyramiddrop,hu2024illava,zhang2024sparsevlm,ye2024atp} compute the attention value by averaging all the attention heads, and could focus on the wrong visual area with hallucination~\cite{zhang2024seeingclearlylayertwo,jiang2024devilsmiddlelayerslarge,yang2025mitigating}, as shown in \figurename~\ref{fig:figattention}(a). On the contrary, we find that not all the self-attention heads in LLMs are relevant to accurately grounding the text tokens to visual tokens.
%Different from existing methods~\cite{chen2025image,xing2024pyramiddrop,hu2024illava,zhang2024sparsevlm,ye2024atp} that compute the attention value by averaging from all attention heads, we find that not all self-attention heads in LLMs are relevant for accurately grounding the text tokens to the visual tokens. As shown in \figurename~\ref{fig:figattention}(a), some attention heads focus on the wrong visual area with hallucination. Inspired by~\cite{zhang2024seeingclearlylayertwo,jiang2024devilsmiddlelayerslarge,yang2025mitigating}, 
%we adopt the magnitude of the attention value between the text token $t$ and all visual tokens to measure the quality of each attention head, denoted as Visual Attention Value (VAV) for the $h$-th head.
% \vspace{-0.1cm}
% \begin{equation}
%     \operatorname{VAV}_h \triangleq \sum_{i=1}^{N} a_{i,h}
%     \vspace{-0.1cm}
% \end{equation}
% where $a_{i,h}$ is the attention value of the $h$-th attention head.
\figurename~\ref{fig:figattention}(b) shows that $\operatorname{VAV}_h$ varies a lot across attention heads, and most heads are close to zero, which are irrelevant. Since heads with higher value are more reliable, we sort and retain the top-$k$ most significant heads as $\boldsymbol{I}=\arg\max(\operatorname{VAV}_1,...,\operatorname{VAV}_H)[:k]$, and derive the importance criteria as $a'_i=\sum_{h = 1}^{k} a_{i,\boldsymbol{I}(h)}$ to identify redundant visual tokens. 
% Then the visual tokens are sorted and the most important $K$ tokens would be retained, while the rest are pruned out in successive layers.
Then the visual tokens are sorted and tokens with large importance values are retained while the rest are pruned out in successive layers.

\begin{table*}[!t]
  \centering
  \resizebox{\linewidth}{!}{
  \begin{tabular}{l l p{6mm}p{6mm}p{6mm}p{6mm} | p{5mm}p{5mm}p{6mm}| p{5mm}p{5.5mm}p{6mm}p{6.5mm} | p{5mm} p{5mm} p{5mm} p{5.5mm}}
  % \begin{tabular}{l l cccc | ccc| ccc | ccccc}
  \toprule
  \multicolumn{6}{c|}{} & \multicolumn{3}{c|}{\emph{Knowledge}} & 
  \multicolumn{4}{c|}{\emph{General}} & \multicolumn{4}{c}{\emph{OCR and Chart}}\\
  % \cmidrule{7-15}
  Method & LLM & \emph{Eff. Res.}  & \emph{Vis. Tok.} &PT & SFT & SQA & AI2D& OK VQA & GQA &POPE &SEED&MMB&Text VQA & Doc VQA& Chart QA& OCR Bench\\
  \midrule
  InstructBLIP~\cite{dai2023instructblip}&Vic.7B&224&32&129M&1.2M&60.5&-&-&49.2&-&53.4&-&50.1&-&-&291\\
  Qwen-VL~\cite{Qwen-VL}&Qwen&448&256&1.4B&50M&67.1&-&-&59.3&-&56.3&38.2&61.5&65.1&-&-\\
  Qwen-VL-Chat~\cite{Qwen-VL}&Qwen&448&256&1.4B&50M&68.2&-&-&57.5&-&58.2&60.6&63.8&62.6&-&488\\
  mPLUG-Owl2~\cite{ye2024mplug}&LLa-2&448&1024&400M&1.2M&68.7&-&-&56.1&-&57.8&64.5&54.3&-&-&366\\
  MM1~\cite{mckinzie2024mm1}&-&1344&720&3B&1.5M&62.3&-&-&-&87.4&65.6&-&68.2&68.4&-&-\\
  LLaVA-Next~\cite{liu2024llavanext}& Vic.7B &672 & 2880 & {558K} & {765K}&70.1&66.6&44.3&64.2&86.5&70.2&67.4&64.9&70.0&64.0&532\\
  \midrule
  Mini-Gemini~\cite{li2024minigeminiminingpotentialmultimodality}&Vic.7B &1536&2880&1.5M&1.5M&-&-&-&-&86.8&-&65.8&68.4&65.0&-&456\\
  SPHINX-2k~\cite{lin2023sphinx}&Vic.13B&762&2890&-&-&70.6&-&-&63.1&87.2&-&65.9&61.2&-&-&-\\
  Pyramiddrop~\cite{xing2024pyramiddrop}& Vic.7B &672 & 1169 & {558K} & {1.0M}&-&-&-&63.7&86.5&69.5&68.1&67.7&66.6&63.0&-\\
  LLaVA-HR~\cite{luo2024feast}&LLa-2&1024&1024&558K&1.2M&65.1&-&-&64.2&87.6&64.2&-&67.1&-&-&-\\
  EAGLE~\cite{shi2024eagleexploringdesignspace}&Vic.7B&1024&1024&558K&1.8M&70.8&-&-&64.9&88.7&73.4&68.1&71.9&-&67.9&544\\
  EAGLE$\dagger$~\cite{shi2024eagleexploringdesignspace}&Vic.7B&1024&1024&558K&1.8M&71.0&72.2&58.8&64.8&88.4&73.5&68.0&71.7&73.2&67.4&538\\
  TokenPacker~\cite{li2024tokenpacker}&Vic.7B &1088&954&1.5M&1.5M&-&-&-&-&88.3&-&67.4&68.0&60.2&-&452\\ Mousi~\cite{fan2024mousipolyvisualexpertvisionlanguagemodels}&Vic.7B&336&576&1.2M&1.6M&69.0&-&-&63.6&86.5&67.5&67.4&58.4&-&-&-\\
  
  DeepStack-V~\cite{meng2024deepstack}&Vic.7B &672&576&558K&665K&-&-&-&64.1&87.6&62.3&-&63.5&41.0&21.0&-\\
  Brave-X5~\cite{kar2024brave}&FlanT5&224&160&100M&-&-&-&-&52.7&87.6&-&-&-&-&-&-\\
  \midrule
  % EAGLE$\dagger$~\cite{shi2024eagleexploringdesignspace}&Vic.7B&1024&1024&558K&1.0M&69.8&67.9&58.8&65.4&87.2&71.3&-&71.4&72.2&67.7&521\\
  % \rowcolor{custom_blue}
  % Ours&Vic.7B &1024&312*&558K&1.0M&70.0&66.8&59.0&64.0&86.4&71.0&69.4&71.3&70.1&65.6&552\\
  % \rowcolor{custom_blue}
  % Ours&Vic.7B &1024&242*&558K&1.0M&69.9&66.7&59.0&63.6&86.2&70.7&69.6&70.7&69.0&65.1&550\\
  % \rowcolor{custom_blue}
  % Ours&Vic.7B &1024&312*&558K&1.8M&71.3&73.7&59.0&63.8&87.8&73.0&69.6&72.0&72.9&66.1&559\\
  \rowcolor{custom_blue}
  \modelname(Ours)&Vic.7B &1024&242*&558K&1.8M&71.4&73.4&58.6&63.5&87.9&72.8&69.8&71.1&71.4&65.6&533\\
  \rowcolor{custom_blue}\modelname(Ours)&LLa-3 &1024&242*&558K&1.8M&77.1&76.9&59.0&63.5&88.2&74.0&72.5&70.8&72.0&68.0&542\\
  % \rowcolor{custom_blue}
  % Ours&Vic.7B &1024&126*&558K&1.0M&69.7&66.4&58.4&63.5&86.9&69.8&-&70.1&66.7&64.6&523\\
  
  % Ours&LLa-3 &1024&242*&558K&1.8M&73.4&71.4&58.6&63.5&87.9&72.8&71.0&71.1&71.4&65.6&560\\
  \rowcolor{custom_blue}
  \modelname(Ours)&Vic.7B&1024&126*&558K&1.8M&70.7&73.9&58.6&62.9&87.6&71.7&69.7&69.4&67.9&64.3&517\\
  \rowcolor{custom_blue}
  \modelname(Ours)&LLa-3&1024&126*&558K&1.8M&78.2&77.1&59.0&64.3&87.8&73.1&72.5&69.0&68.3&66.5&525\\
  \bottomrule
  \end{tabular}
  }
  \vspace{-3mm}
  \caption{Comparison with other MLMMs on 11 benchmarks. \emph{Eff. Res.} indicates the effective image resolution. \emph{Vis. Tok.} indicates the number of visual tokens used in LLMs. “Vic.”, “LLa-2” and “LLa-3” refer to Vicuna, LLaMA-2 and LLaMA-3. $\dagger$ denotes that we reproduce the results with the officially released code and training data (w/o pre-alignment stage).
  Token with (*) means the retained token count is not pre-defined and adaptively determined, which is calculated by averaging across all benchmarks during inference.
  }
  \label{tab1:mainresults}
 \vspace{-3mm}
\end{table*}

%\noindent \textbf{Instance-adaptive Token Retention Number Guided by Visual Attention Value (VAV).} We illustrate the VAV of top-$k$ most significant attention heads on different datasets. 

% As shown in Fig.~\ref{fig:figattention} (c), the value varies and aligns with the complexity of the instances for different datasets, i.e., AI2D with coarse-grained general comprehensive exhibits low value while DocVQA requiring fine-grained OCR recognition displays high value. Thus, we use the VAV to quantify the extent of the text token's interaction with visual information: a higher VAV value indicating a greater contribution from image tokens during the generation.
% \vspace{-0.2cm}
% \setlength{\FrameRule}{1pt} % 设置边框线的宽度
% \setlength{\FrameSep}{2pt} % 设置边框线与内容之间的间距
% \setlength{\leftmargin}{0pt} % 将左边框与左边距对齐
% \begin{framed}
% \noindent \textbf{Finding 4}: The distribution of Visual Attention Value varies a lot for different datasets, and is closely related to the complexity of input instances.
% \end{framed}
% \vspace{-0.2cm}

\noindent \textbf{Instance-adaptive Token Retention.} We find the VAV of top-$k$ most significant attention heads correlates with the complexity of varying inputs, and thus propose to dynamically adjust the retained token numbers based on VAV.

\noindent \emph{\textbf{Finding 5}: The distribution of Visual Attention Value varies a lot for different datasets, and is closely related to the complexity of input instances.}

\figurename~\ref{fig:figattention}(c) exhibits a clear correlation between the VAV and instance complexity across datasets. Specifically, AI2D characterized by coarse-grained general comprehension exhibits low value, while DocVQA demanding fine-grained OCR recognition shows high VAV. Actually, the VAV quantifies the degree to which text tokens interact with visual information, with higher values signifying a greater contribution of image tokens during generation. Thus, we define the visual contribution level as the sum of VAV of top-$k$ heads $\sum_{h=1}^k\sum_{i=1}^Na_{i,\boldsymbol{I}(h)}$. This increased contribution, in turn, necessitates retaining a larger number of visual tokens during generation. 
Consequently, the visual contribution level determines the number of visual tokens to retain as: 
\vspace{-0.1cm}
\begin{equation}
    K=\lambda\cdot\sum_{h=1}^k\sum_{i=1}^Na_{i,\boldsymbol{I}(h)}
    \vspace{-0.07cm}
\end{equation}
where $\lambda$ is a constant for scaling. Besides, since the redundancy of visual tokens gradually increases as the layer deepens~\cite{chen2025image}, we progressively compress tokens at three phases.

\begin{table*}[!t]
  \centering
  \resizebox{\linewidth}{!}{
  \begin{tabular}{l l ll| p{4.5mm}p{4.5mm}p{6mm}| p{5mm}p{5.5mm}p{6mm}p{6.5mm} | p{5mm} p{5.2mm} p{5.2mm} p{7mm}|c}
  % \begin{tabular}{l l cccc | ccc| ccc | ccccc}
  \toprule
  \multicolumn{4}{c|}{} & \multicolumn{3}{c|}{\emph{Knowledge}} & 
  \multicolumn{4}{c|}{\emph{General}} & \multicolumn{4}{c|}{\emph{OCR and Chart}}\\
  % \cmidrule{7-15}
  Method & \emph{Vis. Tok.} & TFOLPS  & Throughput& SQA & AI2D& OK VQA & GQA &POPE &SEED&MMB&Text VQA & Doc VQA& Chart QA& OCR Bench&Avg.\\
  \midrule EAGLE$\dagger$~\cite{shi2024eagleexploringdesignspace}&1024&26.21&0.81&71.0&72.2&58.8&64.8&88.4&73.5&68.0&71.7&73.2&67.4&538&69.3\\
  FastV~\cite{chen2025image}&256&16.83($\downarrow$36\%)&0.92($\uparrow$14\%)&70.5&72.5&57.9&61.8&86.4&69.3&68.1&70.5&54.2&60.4&431&64.9\\
  VTW~\cite{lin2024boosting}&256&15.88($\downarrow$39\%)&0.96($\uparrow$19\%)&59.2&66.6&52.8&52.3&77.8&54.3&63.9&57.1&41.0&51.7&338&55.5\\
  Pdrop~\cite{xing2024pyramiddrop}&256&16.49($\downarrow$37\%)&0.98($\uparrow$21\%)&70.6&72.6&57.8&59.5&86.3&69.3&67.8&68.5&50.4&56.1&392&63.5\\
  SparseVLM~\cite{zhang2024sparsevlm}&256&17.89($\downarrow$32\%)&0.89($\uparrow$10\%)&70.7&72.2&58.2&59.8&86.5&70.2&67.9&70.4&51.9&56.6&383&63.9\\
  PixelShuffle~\cite{chen2024internvlscalingvisionfoundation}&256&16.31($\downarrow$38\%)&1.02($\uparrow$26\%)&69.4&69.7&55.0&62.0&87.7&71.4&68.2&66.9&56.4&60.6&402&64.3\\
  DeCo~\cite{yao2024deco}&256&16.27($\downarrow$38\%)&1.08($\uparrow$33\%)&70.1&70.3&54.3&61.9&87.4&71.3&68.0&66.3&56.3&60.3&389&64.1\\
  % \midrule
  % \rowcolor{custom_blue}
  \modelname(Ours)&242*&13.42($\downarrow$49\%)&1.18($\uparrow$46\%)&71.4&73.4&58.6&63.5&87.9&72.8&69.8&71.1&71.4&65.6&533&69.0\\
  % \rowcolor{custom_blue}
  % Ours&126*&11.40($\downarrow$57\%)&1.24($\uparrow$53\%)&70.7&73.9&58.6&62.9&87.6&71.7&69.9&69.4&67.9&64.3&517&68.1\\
  \bottomrule
  \end{tabular}
  }
  \vspace{-3mm}
  \caption{Comparison with efficient MLLMs methods on 11 benchmarks. EAGLE and our model employ the same training data. $^\dagger$ implies reproducing the results with officially released code on Ascend 910B. Other compared methods are implemented based on EAGLE.
  }
  \label{tab1:mainefficient}
  \vspace{1.6mm}
  % \end{table*}
  % \begin{table*}[!t]
  \centering
  \addtolength{\tabcolsep}{0.7pt}
  {
  \scalebox{0.92}{
  \begin{tabular}{l l | p{4.5mm}p{5mm}p{5mm}p{6.5mm}| p{4.5mm}p{5.2mm}p{5.2mm}p{5.2mm}p{5.8mm} | p{4.8mm}p{5mm} p{5.8mm} p{5.2mm} p{7.2mm}}
  % \begin{tabular}{l l cccc | ccc| ccc | ccccc}
  \toprule
  \multicolumn{2}{c|}{} & \multicolumn{4}{c|}{\emph{Knowledge}} & 
  \multicolumn{5}{c|}{\emph{General}} & \multicolumn{5}{c}{\emph{OCR and Chart}}\\
  % \cmidrule{7-15}
  Method & \emph{Vis. Tok.}& Avg & SQA& MM MU & AI2D &Avg&MME&MMB &SEED&GQA&Avg&Chart QA& OCR Bench&Text VQA & Doc VQA\\
  \midrule
  {MGM-HD}&\quad 2880 & 62.0 & 75.1 & 37.3 & 73.5 & 72.7 & {1606} & 72.7 & 73.2 & 64.5 & 62.9 & 59.1 & 47.7 & 70.2 & 74.6 \\
{Cambrian-1}&\quad 576&65.4& 80.4 & 42.7 & 73.0 & 73.1 & 1547 & 75.9 & 74.7 & 64.6 & 71.3 & 73.3 & 62.4 & 71.7 & 77.8 \\
\modelname(Ours)& \quad 324*&65.4&77.3&42.2&76.8&73.5&1545&78.4&74.4&64.1&74.6&78.5&62.5&	73.8&83.7\\
\bottomrule
\end{tabular}}
}
\vspace{-3.5mm}
\caption{{Results using the same training data and vision encoders as {Cambrian-1} and LLM is Llama3-8B.}
}
\vspace{-3.9mm} 
\label{tab:vqa-results-cambrain}
\end{table*}
\vspace{-0.12cm}
\section{Experiments}
\vspace{-0.05cm}
\subsection{Experimental Settings}
\vspace{-0.05cm}
\textbf{Implementation Details.} We adopt EAGLE~\cite{shi2024eagleexploringdesignspace} as the base settings, where Vicuna-v1.5-7B~\cite{vicuna} and Llama3-8B~\cite{touvron2023llama} are used as the LLM, and four vision encoders: CLIP~\cite{radford2021learning}, ConvNeXt~\cite{liu2022convnet}, Pix2Struct~\cite{lee2023pix2struct} and EVA-02~\cite{sun2023eva} are adopted. For the pre-training stage, we adopt the token discarding strategy in multi-vision encoding and use the same pre-training data as LLaVA-1.5~\cite{liu2023improvedllava} with 558k image-text pairs, where the whole model is frozen and only the projector is updated. For the supervised fine-tuning, based on the pre-trained projector for alignment, we incorporate cooperative pruning in multi-vision fusion with 576 tokens retained and employ the data recipe in~\cite{chen2024open} with 1M image-text pairs as the base setup to finetune the whole model, and also employ EAGLE-1.8M~\cite{shi2024eagleexploringdesignspace} data recipe as the advanced setup. Finally, the proposed instance-adaptive Text-guided Pruning is conducted in a training-free manner with different token budget configurations for simplicity. More details are in the supplementary material.

% The projection layer is an MLP to project the visual embedding into the text embedding space. 
% For the pre-training stage, we use the same pre-training data as LLaVA-1.5~\cite{liu2023improvedllava} with 595k image-text pairs. For the supervised fine-tuning, we employ the data recipe in Open-LLaVA-NeXT~\cite{chen2024open} with 1M image-text pairs.

% \begin{table*}[!t]
% \small
% \centering
% \renewcommand{\arraystretch}{1} 
% \addtolength{\tabcolsep}{-0.75pt}
% \scalebox{0.95}{
% \begin{tabular}{l|l|cccc|ccccc|ccccc}
% \toprule
% {Model}& {$\#$Vis. Tok.}& \multicolumn{4}{c}{Knowledge} & \multicolumn{5}{c}{General} & \multicolumn{5}{c}{OCR and Chart} \\
% \hline
% &&  \rotatebox{90}{Avg} & \rotatebox{90}{SQA$^{\mathrm{I}}$} & \rotatebox{90}{MMMU}  & \rotatebox{90}{AI2D} & \rotatebox{90}{Avg} & \rotatebox{90}{MME} & \rotatebox{90}{MMBench} & \rotatebox{90}{SEED} & \rotatebox{90}{GQA} & \rotatebox{90}{Avg} & \rotatebox{90}{ChartQA} & \rotatebox{90}{OCRBench} & \rotatebox{90}{TextVQA} & \rotatebox{90}{DocVQA}\\
% \hline
% {MGM-HD}&\quad 2880 & 62.0 & 75.1 & 37.3 & 73.5 & 72.7 & {1606} & 72.7 & 73.2 & 64.5 & 62.9 & 59.1 & 47.7 & 70.2 & 74.6 \\
% {Cambrian-1}&\quad 576&65.4& 80.4 & 42.7 & 73.0 & 73.1 & 1547 & 75.9 & 74.7 & 64.6 & 71.3 & 73.3 & 62.4 & 71.7 & 77.8 \\
% \modelname(Ours)& \quad 324*&65.4&77.3&42.2&76.8&73.5&1545&78.4&74.4&64.1&74.6&78.5&62.5&	73.8&83.7\\
% \bottomrule
% \end{tabular}
% }
% \vspace{-3mm}
% \caption{{Results using the same training data and vision experts as {Cambrian-1} and LLM is Llama3-8B.} .
% }
% \vspace{-3mm} 
% \label{tab:vqa-results-cambrain}
% \end{table*}

\noindent \textbf{Evaluation Datasets.} We evaluate the model on existing diverse multimodal benchmarks including SEEDBench~\cite{li2023seed}, POPE~\cite{li2023obj_Hallucination}, TextVQA~\cite{singh2019towards}, ChartQA~\cite{masry2022chartqa}, DocVQA~\cite{mathew2021docvqa}, GQA~\cite{hudson2019gqa}, ScienceQA~\cite{lu2022learn}, AI2D~\cite{kembhavi2016diagram}, OCRBench~\cite{liu2023hidden}, OKVQA~\cite{marino2019ok}, MME~\cite{fu2023mme} and MMBench~\cite{liu2024mmbench}.

% \begin{figure}[!t]
% \begin{minipage}[c]{.483\linewidth}
%     \includegraphics[width=0.81\linewidth,height=0.63\linewidth]{pic/throughoutacc1.png}
%     \vspace{-0.37cm}
%     \caption{\small Average accuracy and throughput of original model, pruning with our strategy, attention or similarity for all stages.}
%     \label{fig:abvis}
% \end{minipage}
% \hspace{-1mm}
% \begin{minipage}[c]{.512\linewidth}
%     \centering
%     \footnotesize
%     % \vspace{-0.05cm}
%     \renewcommand{\arraystretch}{1.2}
%     \setlength{\tabcolsep}{0.1mm}{
%     \begin{tabular}{c|ccc}
%     \toprule
%     Strategy& \textbf{Knowledge} & \textbf{General} & \textbf{OCR}  \\
%     \midrule
%     Rank-based&{65.5}&{74.0}&{65.4}\\
%     Average&65.1&73.8&63.0\\
%     Rank-reverse&64.8&72.5&61.2\\
%     \bottomrule
%     \end{tabular}}
%   \vspace{0.15cm}
%     \captionof{table}{\small Ablation study on the allocation strategies of retained visual token number budget for different vision experts.}
%     \label{tab:abtokenbudget}
% \end{minipage}
% \end{figure}

\vspace{-0.05cm}
\subsection{Main Results}
\vspace{-0.05cm}
\textbf{Comparison with Leading MLLMs.} As shown in Table~\ref{tab1:mainresults}, we present the performance of our method on 11 vision-language benchmarks with two visual token count configurations, i.e., around 242 and 126 tokens. Compared with existing MLLMs, \modelname consistently achieves superior performance, outperforming LLaVA-Next with less than 10\% of the tokens in 10 out of 11 benchmarks. Compared with existing training-based token pruning approaches such as Pyramiddrop, TokenPacker and DeepStack-V, our method demonstrates superior performance with 60\%-80\% fewer tokens, particularly in more challenging fine-grained OCR recognition, e.g., outperforms by 2.4\%, 8.8\% and 28\% on DocVQA, respectively. Compared with existing multi-expert MLLMs, \modelname significantly outperforms Mousi and Brave-X5 with fewer retained visual tokens and less training data, demonstrating the superiority of our token pruning strategy over MLP and Q-former for integrating tokens from multi-vision encoders. Besides, by equipped with a more advanced LLM, the performance of our model can be further improved, outperforming EAGLE on 7 out of 11 benchmarks with less than 25\% of visual tokens. Moreover, even with 126 visual tokens, our method achieves satisfactory performance and consistently performs better than Mini-Gemini, exhibiting the effectiveness in achieving a superior accuracy-efficiency trade-off.

\begin{figure}[!t]
\centering
\renewcommand{\baselinestretch}{1.0}
\centering
\subfloat[Acc. vs. Throughput.]{\includegraphics[width=0.49\linewidth,height=0.38\linewidth]{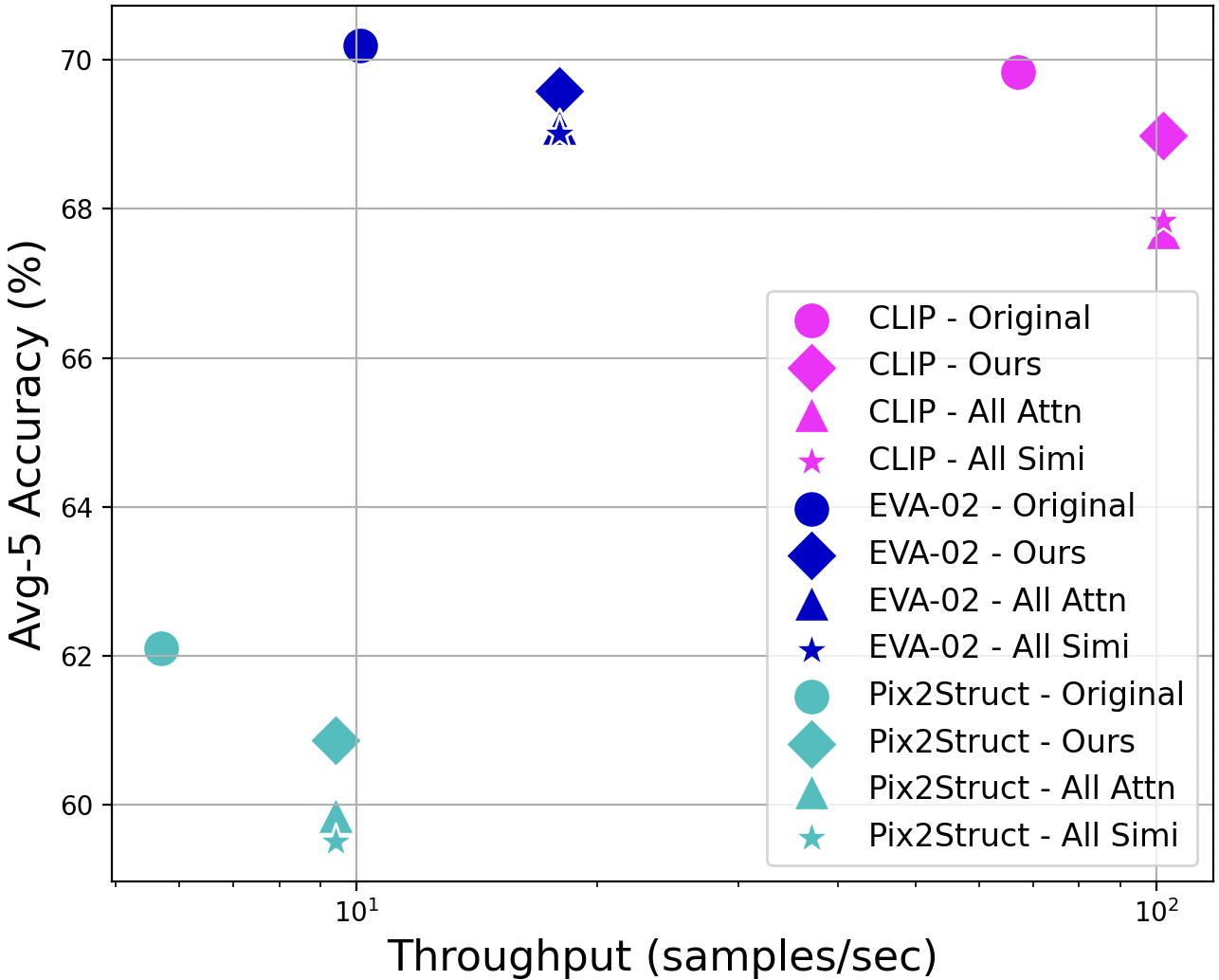}}
\hspace{2pt}
\subfloat[Ablation on allocation strategies.]{\includegraphics[width=0.49\linewidth,height=0.38\linewidth]{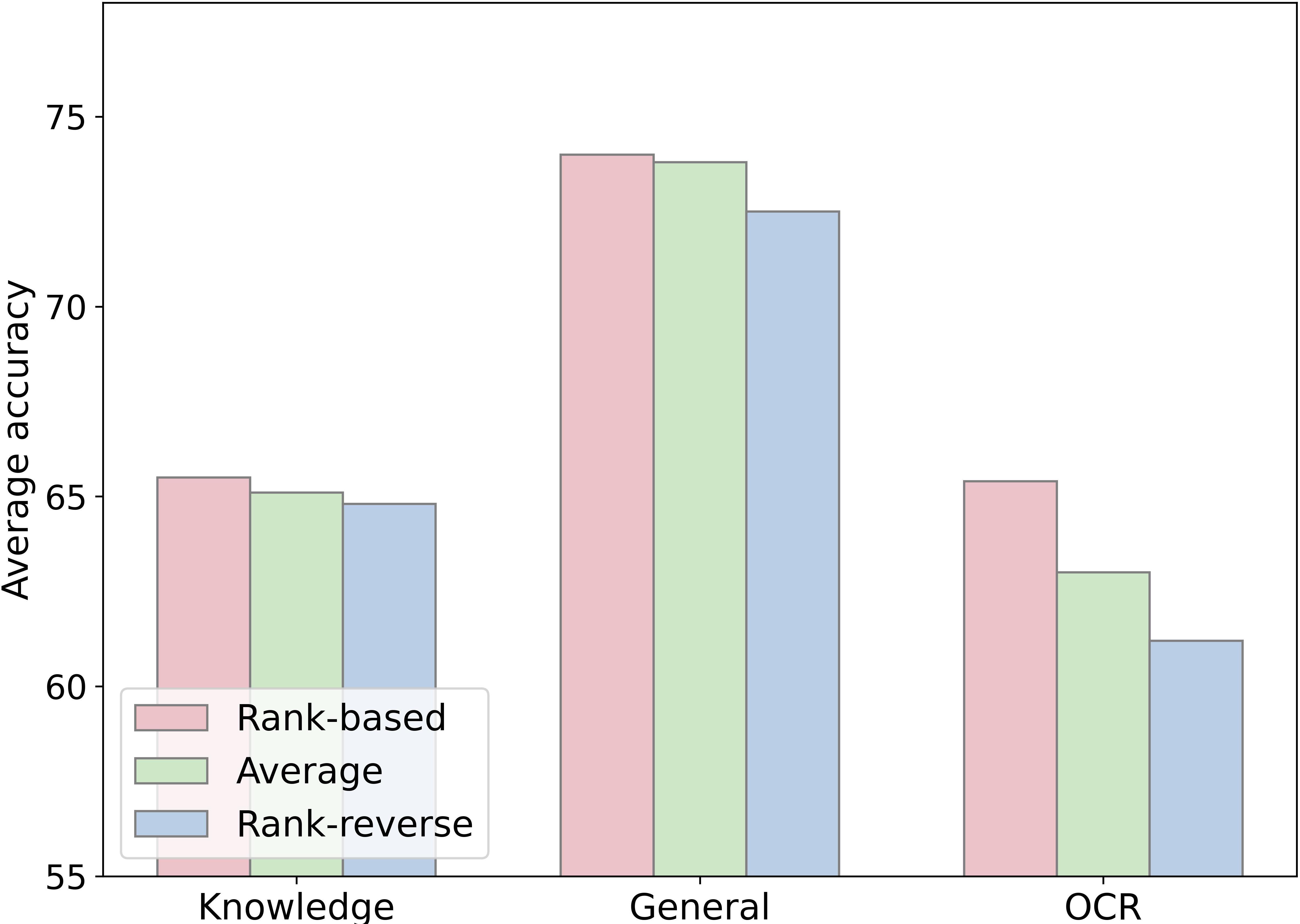}}
\vspace{-10pt}
\caption{(a) Accuracy and throughput of full model, pruning with our strategy, attention or similarity for all layers. (b) Ablation on allocation strategies of token sparsity ratio for each encoder.}\label{fig:sec32ab}
\vspace{-0.5cm}
\end{figure}

\noindent \textbf{Comparison with Existing Efficient MLLM Methods.} Our basic setup follows EAGLE~\cite{shi2024eagleexploringdesignspace} with the same vision encoders and training data, except that we do not involve the pre-alignment stage for simplicity. Thus, we compare with existing efficient MLLM methods based on EAGLE. All the results are tested on Ascend 910B. Table~\ref{tab1:mainefficient} shows that \modelname could reduce 49\% TFLOPS and increase 46\% throughput with only a 0.3\% performance drop on average, showing obvious advantages over existing methods. LLM-based pruning methods such as FastV, Pdrop and SparseVLM ignore the acceleration of vision encoding process, which also takes up a considerable time as the image resolution increases. Besides, a pre-defined pruning ratio for all datasets fails to achieve a promising trade-off between accuracy and efficiency. Moreover, despite satisfactory performance on general benchmarks, training-based methods Deco and PixelShuffle suffer from a significant performance drop on OCR benchmarks, since they fail to achieve the effective instance-adaptive pruning ratio, where OCR tasks demand much more tokens than general tasks.
\begin{table}[!t]
    \small
    \centering
    \renewcommand{\arraystretch}{0.97} 
    \addtolength{\tabcolsep}{-3.8pt}
    \scalebox{0.97}{
    \begin{tabular}{ccc|ccc|c}
    \hline
    Fusion&Pruning&\#Tok.& \textbf{Knowledge} & \textbf{General} & \textbf{OCR} &\textbf{Avg.} \\
    \hline
    Pre-proj&-&1193&64.4&73.7&64.5&67.5\\
    Post-proj&-&1193&65.5&74.0&65.4&68.3\\  
    \hline
    Post-proj&Separate&576&64.8&73.4&62.4&66.9\\
    Post-proj&Random&576&63.4&72.2&48.5&61.4\\
    Post-proj&MLP&576&63.8&72.5&55.9&64.1\\
    Post-proj&PixelShuffle&576&64.1&72.5&55.1&63.9\\
    Post-proj&Resampler&576&61.1&70.6&49.2&60.3\\
    Post-proj&Ours&576&65.4&74.0&65.5&68.3\\
    \hline
    \end{tabular}
    }
    \vspace{-3mm}
    \caption{Ablation study on the fusion strategy, i.e., pre-projection and post-projection fusion, and pruning strategy.}
    \vspace{-5.3mm} 
    \label{tab:abtokenfusion}
\end{table}

\noindent \textbf{Extension with Different Vision Encoders.} To validate the generalizability of our method, we adopt another group of vision encoders as SigLIP, ConvNeXt, CLIP and DINOv2 following Cambrian-1~\cite{tong2024cambrian1fullyopenvisioncentric}. 
Table~\ref{tab:vqa-results-cambrain} shows that \modelname outperforms MGM-HD and Cambrian-1 with fewer visual tokens, especially outperforming by 11.7\% and 3.3\% accuracy on OCR tasks, exhibiting the effectiveness of our strategy over spatial vision aggregator~\cite{tong2024cambrian1fullyopenvisioncentric} for adaptively compressing visual tokens from multi-vision encoders.
\begin{table*}[!t]
  \centering
  \small 
  \renewcommand{\arraystretch}{0.8} 
  \addtolength{\tabcolsep}{1pt}
  \begin{tabular}{l|c|c|c|c|cccc}
    \toprule[1pt]
    Pruning &Attn-head& Avg.  & Pruning & Avg.  & \multirow{2}{*}{\textbf{Knowledge}} & \multirow{2}{*}{\textbf{General}} & \multirow{2}{*}{\textbf{OCR}} & \multirow{2}{*}{\textbf{Avg.}}  \\
     Strategy&Filtering & Tokens  & Layer Indexes & Retained Tokens &  &  &  &   \\
    \midrule[1pt]
    Upper Bound & - & 576 & - & - & 65.4 & 74.0 & 65.5 & 68.3     \\
    \midrule
    \multirow{3}{*}{Pre-defined Ratio}&  \XSolidBrush& 297  & [1]        & [288]         & 65.0 & 73.2 & 59.1 & 65.8     \\
    &\Checkmark& 297 & [1] & [288] & 65.1 & 73.3 & 59.9 & 66.1     \\
  &\Checkmark& 242 & [4, 12, 20] & [390, 172, 78] & 65.0 & 73.3 & 62.4 & 66.9     \\
    \midrule
    \multirow{3}{*}{{Our adaptive Ratio}}& \multirow{3}{*}{\Checkmark}   & 312\textsuperscript{*} & \multirow{3}{*}{[4, 12, 20]}  & [{525\textsuperscript{*}}, {252\textsuperscript{*}}, {122\textsuperscript{*}}] & {65.2} & {73.8} & {65.6} &{68.2}     \\
     && 242\textsuperscript{*} &  & [{396\textsuperscript{*}}, {170\textsuperscript{*}}, {76\textsuperscript{*}}] & 65.2&73.5&65.0&67.9  \\
     && 126\textsuperscript{*} &  & [{226\textsuperscript{*}}, {80\textsuperscript{*}}, {36\textsuperscript{*}}] & {64.8} & 73.4 & 63.4 & 67.2     \\
    \bottomrule[1pt]
  \end{tabular}
  \vspace{-2.8mm} 
  \caption{Comparison of pre-defined pruning ratio and our instance-adaptive ratio, w or w/o top-$k$ attention head filtering under the same training-free setting.
  Pruning Layer Indexes specify the LLM layer indexes at which visual tokens are pruned, starting from 0 and occurring prior to input. Token with (*) means the retained token count is not pre-defined and is calculated by averaging across all benchmarks.}
  \vspace{-3.2mm} 
\label{tab:abllmprune}
\end{table*}
% \begin{figure}[!t]
%     \centering
%     \includegraphics[width=0.97\linewidth,height=0.42\linewidth]{pic/multiencodernumber.png}
%     \vspace{-3mm}
%     \caption{Contribution of vision encoders for POPE and DocVQA.}
%     \label{fig:multiencodernumber}
%     \vspace{-3mm}
% \end{figure}

\vspace{-0.08cm}
\subsection{Ablation Study} 
\vspace{-0.05cm}
Experiments in this subsection are with 558K data for pretraining and 1M data for SFT. Evaluation results are summarized into Knowledge (SQA, AI2D, OKVQA), General (GQA, POPE, SEED) and OCR (TextVQA, DocVQA, ChartQA, OCRBench) by averaging to save space.
% \begin{table}[!t]
%     \small
%     \centering
%     \renewcommand{\arraystretch}{0.99} 
%     % \addtolength{\tabcolsep}{-1pt}
%     \scalebox{1.0}{
%     \begin{tabular}{c|ccc|c}
%     \hline
%     ~Strategy& ~\textbf{Knowledge}~ & ~\textbf{General}~ & ~\textbf{OCR}~ &~\textbf{Avg.}~ \\
%     \hline
%     Rank-based&{65.5}&{74.0}&{65.4}&{68.3}\\
%     Average&65.1&73.8&63.0&67.3\\
%     Rank-reverse&64.8&72.5&61.2&66.2\\
%     \hline
%     \end{tabular}
%     }
%     \vspace{-3mm}
%     \caption{Ablation study on the assigning strategies of visual token number budget for different vision experts. }
%     \label{tab:abtokenbudget}
% \end{table}

\noindent \textbf{Token Pruning in Multi-vision Encoding.} To identify redundant tokens, we employ the similarity to the average token for the shallow block and attention value in the deep blocks. \figurename~\ref{fig:sec32ab}(a) validates the effectiveness of our strategy over using similarity or attention value for all blocks. Besides, to allocate the token sparsity ratio for each encoder, we compare our strategy with 1) Average: equal token numbers to all encoders and 2) Rank-inverse: fewer tokens to encoders generating higher-ranked features. \figurename~\ref{fig:sec32ab}(b) shows our rank-based strategy achieves the best performance, especially on more challenging OCR tasks, validating that low-rank feature maps contain less information
and should be allocated with fewer token numbers.
\begin{figure}[!t]
    \centering
    \includegraphics[width=0.95\linewidth,height=0.47\linewidth]{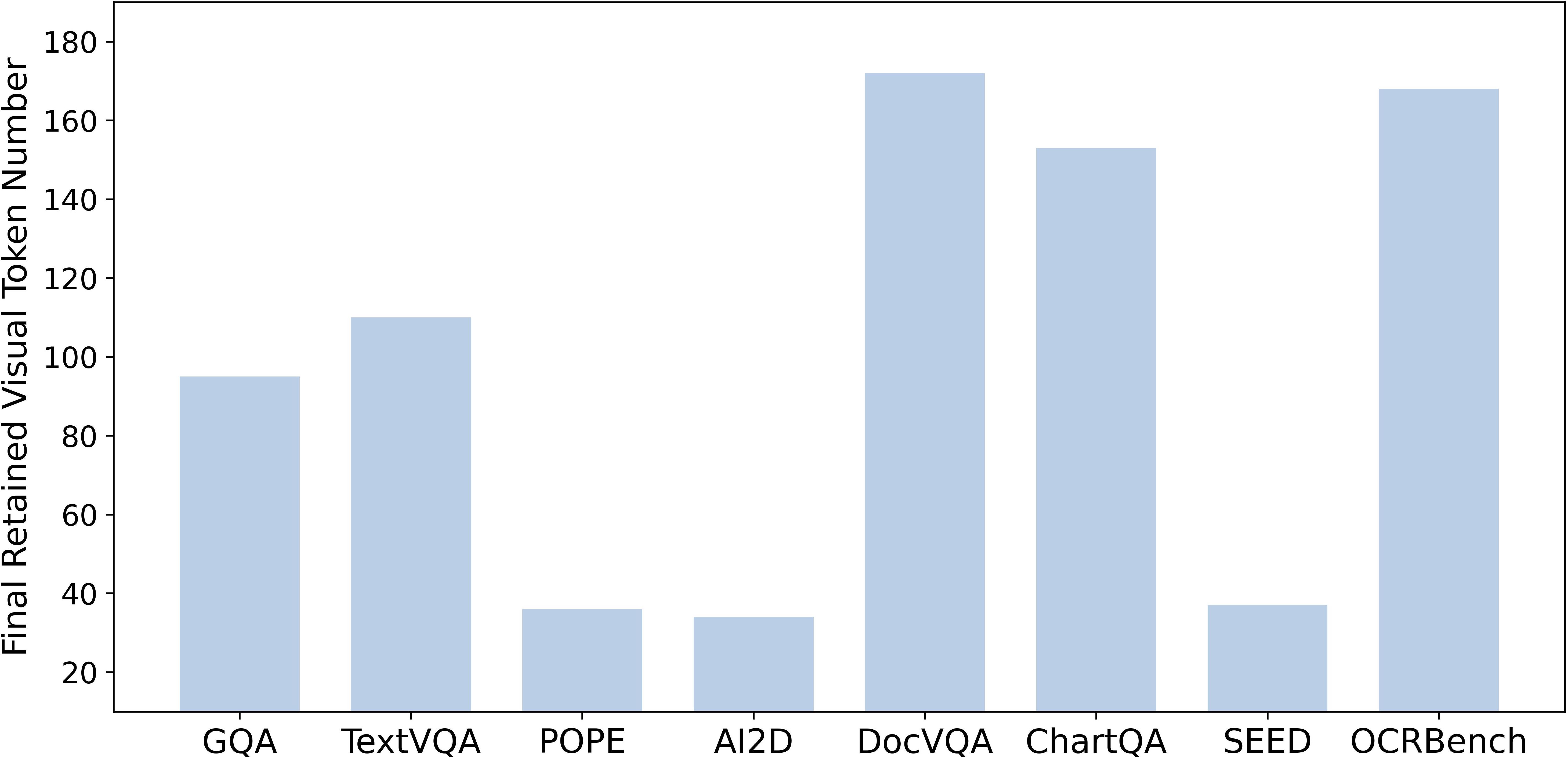}
    \vspace{-2.5mm}
    \caption{Final retained visual token number for different datasets.}
    \label{fig:finaltokennumber}
    \vspace{-4mm}
\end{figure}

% \noindent \textbf{Token Pruning in Multi-vision Fusion.} Previous multi-vision experts based MLLMs usually adopt pre-projection fusion, while we propose a more flexible post-projector fusion strategy that each expert independently adapt visual tokens before fusion, which achieves superior performance as shown in Table~\ref{tab:abtokenfusion}. Moreover, based on the projector for aligning multi-vision experts, we compare our collaborative pruning strategy with 1) Separate pruning: pruning tokens within each expert separately and 2) Random pruning. Table~\ref{tab:abtokenfusion} shows that our strategy significantly outperforms two alternatives, showing the effectiveness of reducing the token redundancy across multi-vision experts, which also exhibits more feature diversity in Fig.~\ref{fig:sec2rank} (b). Compared with parameter-based compression modules like MLP and Resampler, our strategy performs better with the priority of measuring the redundancy with similarity across experts.
\noindent \textbf{Token Pruning in Multi-vision Fusion.} Previous methods usually adopt pre-projection fusion, while we propose a more flexible post-projecton fusion strategy to independently adapt visual tokens of each encoder before fusion, which achieves superior performance in Table~\ref{tab:abtokenfusion}. Moreover, based on the projector for aligning multi-encoders, we compare our pruning strategy with 1) Separate pruning: pruning within each encoder separately and 2) Random pruning. Table~\ref{tab:abtokenfusion} shows that our strategy significantly outperforms two alternatives, showing the effectiveness of discarding the token redundancy across multi-vision encoders, which also exhibits more feature diversity in \figurename~\ref{fig:sec2rank}(b). Our strategy also performs better than parameter-based compression like MLP and Resampler, which require more data for training.

\noindent \textbf{Instance-adaptive Text-guided Token Pruning.} Firstly, Table~\ref{tab:abllmprune} shows our attn-head filtering strategy significantly improves the performance on OCR by 0.8\% accuracy, exhibiting the effectiveness of employing the most relevant attention for accurately measuring the token redundancy. Besides, As shown in Table~\ref{tab:abllmprune}, our strategy consistently outperforms pre-defined fixed pruning rates across all benchmarks with the same average tokens of 242, particularly outperforming by 2.6\% on OCR tasks, which are more complex and require more tokens than general tasks. These results demonstrate that tailoring adaptive pruning strategies to specific instances or tasks for dynamically adjusting the token budget could mitigate performance degradation.

\noindent \textbf{Analysis of Retained Token Number.} \figurename~\ref{fig:finaltokennumber} shows the retained token numbers (pruning ratio) vary a lot across datasets. General tasks like AI2D achieve satisfactory results with few tokens, while OCR recognition requiring detailed visual information benefits from retaining more tokens. This adaptive behavior shows the effectiveness of our method in tailoring token budget for varying task demands.
% \textbf{Contribution of Vision Encoders.} To understand the impact of vision encoders, we evaluate the ratio of retained token numbers for different vision encoders. As shown in Fig.~\ref{fig:multiencodernumber}, EVA and Convnext contribute more to fine-grained POPE, while Pix2Struct and Convnext are more crucial for DocVQA, showing the effectiveness of our method for retaining suitable visual tokens for different tasks.

\noindent\textbf{Contribution of Each Encoder.} We count the ratio of retained tokens from each encoder in supplementary material.

% \noindent \textbf{Instance-adaptive Token Pruning.} To understand the effectiveness of our instance-adaptive token pruning, we count the number of finally retained visual token numbers. As shown in Fig.~\ref{fig:finaltokennumber}, the retained token numbers (i.e., pruning ratio) varies a lot across datasets. General tasks focusing on global understanding achieves satisfactory results with few tokens. While OCR recognition requiring detailed visual information benefits from retaining more tokens. This adaptive behavior validates the effectiveness of \modelname in tailoring token budget to specific task demands.
% For general tasks, quite few tokens can fulfill the global comprehension. While for OCR recognition, much more visual tokens are retained to mitigate performance degradation. These results indicate the effectiveness of our method to achieve adaptively token pruning. 
\vspace{-0.08cm}
\section{Conclusion}
\vspace{-0.03cm}
In this paper, we propose a multi-encoder collaborative token pruning strategy, namely METEOR, to progressively eliminate redundant visual tokens across the encoding, fusion and decoding stages. Based on several interesting findings, we leverage the reliable measure to identify redundant tokens across various stages. Besides, the rank of feature maps is employed as a mathematically grounded measure to allocate the sparse ratio for various encoders. Moreover, we propose an instance-adaptive token pruning strategy to dynamically adjust the pruning ratio for various task demands. Extensive experiments demonstrate the effectiveness of our method. Compared with EAGLE, a typical multi-encoder MLLMs, \modelname reduces 76\% visual tokens with only 0.3\% performance drop, significantly outperforming existing efficient methods.

\section*{Acknowledgements}
This work was supported in part by the National Natural Science Foundation of China under Grant 62431017, Grant 62320106003, Grant U24A20251, Grant 62125109, Grant 62371288, Grant 62301299, Grant 62401357, Grant 62401366,  Grant 62120106007, and in part by the Program of Shanghai Science and Technology Innovation Project under Grant 24BC3200800.

{\small
\bibliographystyle{ieeenat_fullname}
\bibliography{main}

\begin{thebibliography}{84}
\providecommand{\natexlab}[1]{#1}
\providecommand{\url}[1]{\texttt{#1}}
\expandafter\ifx\csname urlstyle\endcsname\relax
  \providecommand{\doi}[1]{doi: #1}\else
  \providecommand{\doi}{doi: \begingroup \urlstyle{rm}\Url}\fi

\bibitem[Arif et~al.(2025)Arif, Yoon, Nikolopoulos, Vandierendonck, John, and Ji]{arif2024hired}
Kazi Hasan~Ibn Arif, JinYi Yoon, Dimitrios~S Nikolopoulos, Hans Vandierendonck, Deepu John, and Bo Ji.
\newblock {HiRED}: Attention-guided token dropping for efficient inference of high-resolution vision-language models.
\newblock In \emph{Proceedings of the 39th AAAI Conference on Artificial Intelligence}, pages 1773--1781, 2025.

\bibitem[Bai et~al.(2023{\natexlab{a}})Bai, Bai, Chu, Cui, Dang, Deng, Fan, Ge, Han, Huang, Hui, Ji, Li, Lin, Lin, Liu, Liu, Lu, Lu, Ma, Men, Ren, Ren, Tan, Tan, Tu, Wang, Wang, Wang, Wu, Xu, Xu, Yang, Yang, Yang, Yang, Yao, Yu, Yuan, Yuan, Zhang, Zhang, Zhang, Zhang, Zhou, Zhou, Zhou, and Zhu]{bai2023qwentechnicalreport}
Jinze Bai, Shuai Bai, Yunfei Chu, Zeyu Cui, Kai Dang, Xiaodong Deng, Yang Fan, Wenbin Ge, Yu Han, Fei Huang, Binyuan Hui, Luo Ji, Mei Li, Junyang Lin, Runji Lin, Dayiheng Liu, Gao Liu, Chengqiang Lu, Keming Lu, Jianxin Ma, Rui Men, Xingzhang Ren, Xuancheng Ren, Chuanqi Tan, Sinan Tan, Jianhong Tu, Peng Wang, Shijie Wang, Wei Wang, Shengguang Wu, Benfeng Xu, Jin Xu, An Yang, Hao Yang, Jian Yang, Shusheng Yang, Yang Yao, Bowen Yu, Hongyi Yuan, Zheng Yuan, Jianwei Zhang, Xingxuan Zhang, Yichang Zhang, Zhenru Zhang, Chang Zhou, Jingren Zhou, Xiaohuan Zhou, and Tianhang Zhu.
\newblock Qwen technical report.
\newblock \emph{arXiv preprint arXiv:2309.16609}, 2023{\natexlab{a}}.

\bibitem[Bai et~al.(2023{\natexlab{b}})Bai, Bai, Yang, Wang, Tan, Wang, Lin, Zhou, and Zhou]{Qwen-VL}
Jinze Bai, Shuai Bai, Shusheng Yang, Shijie Wang, Sinan Tan, Peng Wang, Junyang Lin, Chang Zhou, and Jingren Zhou.
\newblock {Qwen-VL}: A frontier large vision-language model with versatile abilities.
\newblock \emph{arXiv preprint arXiv:2308.12966}, 2023{\natexlab{b}}.

\bibitem[Brown et~al.(2020)Brown, Mann, Ryder, Subbiah, Kaplan, Dhariwal, Neelakantan, Shyam, Sastry, Askell, Agarwal, Herbert-Voss, Krueger, Henighan, Child, Ramesh, Ziegler, Wu, Winter, Hesse, Chen, Sigler, Litwin, Gray, Chess, Clark, Berner, McCandlish, Radford, Sutskever, and Amodei]{brown2020language}
Tom Brown, Benjamin Mann, Nick Ryder, Melanie Subbiah, Jared~D Kaplan, Prafulla Dhariwal, Arvind Neelakantan, Pranav Shyam, Girish Sastry, Amanda Askell, Sandhini Agarwal, Ariel Herbert-Voss, Gretchen Krueger, Tom Henighan, Rewon Child, Aditya Ramesh, Daniel Ziegler, Jeffrey Wu, Clemens Winter, Chris Hesse, Mark Chen, Eric Sigler, Mateusz Litwin, Scott Gray, Benjamin Chess, Jack Clark, Christopher Berner, Sam McCandlish, Alec Radford, Ilya Sutskever, and Dario Amodei.
\newblock Language models are few-shot learners.
\newblock In \emph{Advances in Neural Information Processing Systems 33}, pages 1877--1901, 2020.

\bibitem[Cai et~al.(2025)Cai, Yang, Gao, and Lee]{cai2024matryoshka}
Mu Cai, Jianwei Yang, Jianfeng Gao, and Yong~Jae Lee.
\newblock Matryoshka multimodal models.
\newblock In \emph{The Thirteenth International Conference on Learning Representations}, 2025.

\bibitem[Chen and Xing(2024)]{chen2024open}
Lin Chen and Long Xing.
\newblock {Open-LLaVA-NeXT}: An open-source implementation of {LLaVA-NeXT} series for facilitating the large multi-modal model community.
\newblock \url{https://github.com/xiaoachen98/Open-LLaVA-NeXT}, 2024.

\bibitem[Chen et~al.(2024{\natexlab{a}})Chen, Zhao, Liu, Bai, Lin, Zhou, and Chang]{chen2025image}
Liang Chen, Haozhe Zhao, Tianyu Liu, Shuai Bai, Junyang Lin, Chang Zhou, and Baobao Chang.
\newblock An image is worth 1/2 tokens after layer 2: Plug-and-play inference acceleration for large vision-language models.
\newblock In \emph{Proceedings of the 18th European Conference on Computer Vision (ECCV)}, pages 19--35, 2024{\natexlab{a}}.

\bibitem[Chen et~al.(2024{\natexlab{b}})Chen, Wang, Tian, Ye, Gao, Cui, Tong, Hu, Luo, Ma, Ma, Wang, Dong, Yan, Guo, He, Shi, Jin, Xu, Wang, Wei, Li, Zhang, Zhang, Cai, Wen, Yan, Dou, Lu, Zhu, Lu, Lin, Qiao, Dai, and Wang]{chen2024fargpt4vclosinggap}
Zhe Chen, Weiyun Wang, Hao Tian, Shenglong Ye, Zhangwei Gao, Erfei Cui, Wenwen Tong, Kongzhi Hu, Jiapeng Luo, Zheng Ma, Ji Ma, Jiaqi Wang, Xiaoyi Dong, Hang Yan, Hewei Guo, Conghui He, Botian Shi, Zhenjiang Jin, Chao Xu, Bin Wang, Xingjian Wei, Wei Li, Wenjian Zhang, Bo Zhang, Pinlong Cai, Licheng Wen, Xiangchao Yan, Min Dou, Lewei Lu, Xizhou Zhu, Tong Lu, Dahua Lin, Yu Qiao, Jifeng Dai, and Wenhai Wang.
\newblock How far are we to {GPT-4V}? {Closing} the gap to commercial multimodal models with open-source suites.
\newblock \emph{Science China Information Sciences}, 67\penalty0 (12):\penalty0 220101, 2024{\natexlab{b}}.

\bibitem[Chen et~al.(2024{\natexlab{c}})Chen, Wu, Wang, Su, Chen, Xing, Zhong, Zhang, Zhu, Lu, Li, Luo, Lu, Qiao, and Dai]{chen2024internvlscalingvisionfoundation}
Zhe Chen, Jiannan Wu, Wenhai Wang, Weijie Su, Guo Chen, Sen Xing, Muyan Zhong, Qinglong Zhang, Xizhou Zhu, Lewei Lu, Bin Li, Ping Luo, Tong Lu, Yu Qiao, and Jifeng Dai.
\newblock {Intern VL}: Scaling up vision foundation models and aligning for generic visual-linguistic tasks.
\newblock In \emph{2024 IEEE/CVF Conference on Computer Vision and Pattern Recognition (CVPR)}, pages 24185--24198, 2024{\natexlab{c}}.

\bibitem[Chiang et~al.(2023)Chiang, Li, Lin, Sheng, Wu, Zhang, Zheng, Zhuang, Zhuang, Gonzalez, Stoica, and Xing]{vicuna}
Wei-Lin Chiang, Zhuohan Li, Zi Lin, Ying Sheng, Zhanghao Wu, Hao Zhang, Lianmin Zheng, Siyuan Zhuang, Yonghao Zhuang, Joseph~E. Gonzalez, Ion Stoica, and Eric~P. Xing.
\newblock Vicuna: An open-source chatbot impressing {GPT-4} with 90\%* {ChatGPT} quality.
\newblock \url{https://lmsys.org/blog/2023-03-30-vicuna/}, 2023.

\bibitem[Dai et~al.(2023)Dai, Li, Li, Tiong, Zhao, Wang, Li, Fung, and Hoi]{dai2023instructblip}
Wenliang Dai, Junnan Li, Dongxu Li, Anthony Tiong, Junqi Zhao, Weisheng Wang, Boyang Li, Pascale~N Fung, and Steven Hoi.
\newblock {InstructBLIP}: Towards general-purpose vision-language models with instruction tuning.
\newblock In \emph{Advances in Neural Information Processing Systems 36}, pages 49250--49267, 2023.

\bibitem[Dai et~al.(2019)Dai, Yang, Yang, Carbonell, Le, and Salakhutdinov]{dai2019transformer}
Zihang Dai, Zhilin Yang, Yiming Yang, Jaime Carbonell, Quoc~V Le, and Ruslan Salakhutdinov.
\newblock {Transformer-XL}: Attentive language models beyond a fixed-length context.
\newblock In \emph{Proceedings of the 57th Annual Meeting of the Association for Computational Linguistics}, pages 2978--2988, 2019.

\bibitem[Ding et~al.(2023)Ding, Zhao, Zhang, Qian, Xiong, and Tian]{ding2023prune}
Shuangrui Ding, Peisen Zhao, Xiaopeng Zhang, Rui Qian, Hongkai Xiong, and Qi Tian.
\newblock Prune spatio-temporal tokens by semantic-aware temporal accumulation.
\newblock In \emph{2023 IEEE/CVF Conference on Computer Vision and Pattern Recognition (CVPR)}, pages 16945--16956, 2023.

\bibitem[Fan et~al.(2024)Fan, Ji, Jiang, Li, Jin, Song, Wang, Hong, Chen, Zheng, Zhang, Huang, Zheng, Xi, Zhou, Dou, Ye, Yan, Gui, Zhang, Qiu, Huang, Wu, and Jiang]{fan2024mousipolyvisualexpertvisionlanguagemodels}
Xiaoran Fan, Tao Ji, Changhao Jiang, Shuo Li, Senjie Jin, Sirui Song, Junke Wang, Boyang Hong, Lu Chen, Guodong Zheng, Ming Zhang, Caishuang Huang, Rui Zheng, Zhiheng Xi, Yuhao Zhou, Shihan Dou, Junjie Ye, Hang Yan, Tao Gui, Qi Zhang, Xipeng Qiu, Xuanjing Huang, Zuxuan Wu, and Yu-Gang Jiang.
\newblock {MouSi}: Poly-visual-expert vision-language models.
\newblock \emph{arXiv preprint arXiv:2401.17221}, 2024.

\bibitem[Fu et~al.(2023)Fu, Chen, Shen, Qin, Zhang, Lin, Yang, Zheng, Li, Sun, Wu, and Ji]{fu2023mme}
Chaoyou Fu, Peixian Chen, Yunhang Shen, Yulei Qin, Mengdan Zhang, Xu Lin, Jinrui Yang, Xiawu Zheng, Ke Li, Xing Sun, Yunsheng Wu, and Rongrong Ji.
\newblock {MME}: A comprehensive evaluation benchmark for multimodal large language models.
\newblock \emph{arXiv preprint arXiv:2306.13394}, 2023.

\bibitem[Guo et~al.(2024)Guo, Xu, Yao, Cui, Ni, Ge, Chua, Liu, and Huang]{xu2024llavauhdlmmperceivingaspect}
Zonghao Guo, Ruyi Xu, Yuan Yao, Junbo Cui, Zanlin Ni, Chunjiang Ge, Tat-Seng Chua, Zhiyuan Liu, and Gao Huang.
\newblock Llava-uhd: an lmm perceiving any aspect ratio and high-resolution images.
\newblock In \emph{Proceedings of the 18th European Conference on Computer Vision (ECCV)}, pages 390--406, 2024.

\bibitem[Han et~al.(2024)Han, Liu, Ding, Wang, Chen, Yan, and Huang]{han2024rethinking}
Yuhang Han, Xuyang Liu, Pengxiang Ding, Donglin Wang, Honggang Chen, Qingsen Yan, and Siteng Huang.
\newblock Rethinking token reduction in {MLLMs}: Towards a unified paradigm for training-free acceleration.
\newblock \emph{arXiv preprint arXiv:2411.17686}, 2024.

\bibitem[He et~al.(2024{\natexlab{a}})He, Wei, Xie, and Tian]{he2024incorporatingvisualexpertsresolve}
Xin He, Longhui Wei, Lingxi Xie, and Qi Tian.
\newblock Incorporating visual experts to resolve the information loss in multimodal large language models.
\newblock \emph{arXiv preprint arXiv:2401.03105}, 2024{\natexlab{a}}.

\bibitem[He et~al.(2024{\natexlab{b}})He, Chen, Liu, Shao, Zhou, Zhang, and Zhuang]{he2024zipvl}
Yefei He, Feng Chen, Jing Liu, Wenqi Shao, Hong Zhou, Kaipeng Zhang, and Bohan Zhuang.
\newblock {ZipVL}: Efficient large vision-language models with dynamic token sparsification and {KV} cache compression.
\newblock \emph{arXiv preprint arXiv:2410.08584}, 2024{\natexlab{b}}.

\bibitem[Hu et~al.(2024{\natexlab{a}})Hu, Shang, Wan, and Feng]{hu2024illava}
Lianyu Hu, Fanhua Shang, Liang Wan, and Wei Feng.
\newblock {iLLaVA}: An image is worth fewer than 1/3 input tokens in large multimodal models.
\newblock \emph{arXiv preprint arXiv:2412.06263}, 2024{\natexlab{a}}.

\bibitem[Hu et~al.(2024{\natexlab{b}})Hu, Dou, Li, Kamath, Peng, and Chang]{hu2024matryoshka}
Wenbo Hu, Zi-Yi Dou, Liunian~Harold Li, Amita Kamath, Nanyun Peng, and Kai-Wei Chang.
\newblock Matryoshka query {Transformer} for large vision-language models.
\newblock In \emph{Advances in Neural Information Processing Systems 37}, pages 50168--50188, 2024{\natexlab{b}}.

\bibitem[Huang et~al.(2024)Huang, Zou, Xi, Wang, Xie, and Yu]{huang2025ivtp}
Kai Huang, Hao Zou, Ye Xi, BoChen Wang, Zhen Xie, and Liang Yu.
\newblock {IVTP}: Instruction-guided visual token pruning for large vision-language models.
\newblock In \emph{Proceedings of the 18th European Conference on Computer Vision (ECCV)}, pages 214--230, 2024.

\bibitem[Hudson and Manning(2019)]{hudson2019gqa}
Drew~A Hudson and Christopher~D Manning.
\newblock {GQA}: A new dataset for real-world visual reasoning and compositional question answering.
\newblock In \emph{2019 IEEE/CVF Conference on Computer Vision and Pattern Recognition (CVPR)}, pages 6700--6709, 2019.

\bibitem[Jiang et~al.(2023)Jiang, Liu, Liu, Zhao, Zhang, Gao, Zhang, Li, and Xiong]{jiang2024clipdinovisualencoders}
Dongsheng Jiang, Yuchen Liu, Songlin Liu, Jin'e Zhao, Hao Zhang, Zhen Gao, Xiaopeng Zhang, Jin Li, and Hongkai Xiong.
\newblock From {CLIP} to {DINO}: Visual encoders shout in multi-modal large language models.
\newblock \emph{arXiv preprint arXiv:2310.08825}, 2023.

\bibitem[Jiang et~al.(2024)Jiang, Huang, Liu, Zeng, Li, Cheng, and Xu]{jiang2024fopru}
Lei Jiang, Weizhe Huang, Tongxuan Liu, Yuting Zeng, Jing Li, Lechao Cheng, and Xiaohua Xu.
\newblock {FoPru}: Focal pruning for efficient large vision-language models.
\newblock \emph{arXiv preprint arXiv:2411.14164}, 2024.

\bibitem[Jiang et~al.(2025)Jiang, Chen, Zhu, Luo, Shen, and Yang]{jiang2024devilsmiddlelayerslarge}
Zhangqi Jiang, Junkai Chen, Beier Zhu, Tingjin Luo, Yankun Shen, and Xu Yang.
\newblock Devils in middle layers of large vision-language models: Interpreting, detecting and mitigating object hallucinations via attention lens.
\newblock In \emph{2025 IEEE/CVF Conference on Computer Vision and Pattern Recognition (CVPR)}, pages 25004--25014, 2025.

\bibitem[Kar et~al.(2024)Kar, Tonioni, Poklukar, Kulshrestha, Zamir, and Tombari]{kar2024brave}
O{\u{g}}uzhan~Fatih Kar, Alessio Tonioni, Petra Poklukar, Achin Kulshrestha, Amir Zamir, and Federico Tombari.
\newblock {BRAVE}: Broadening the visual encoding of vision-language models.
\newblock In \emph{Proceedings of the 18th European Conference on Computer Vision (ECCV)}, pages 113--132, 2024.

\bibitem[Kembhavi et~al.(2016)Kembhavi, Salvato, Kolve, Seo, Hajishirzi, and Farhadi]{kembhavi2016diagram}
Aniruddha Kembhavi, Mike Salvato, Eric Kolve, Minjoon Seo, Hannaneh Hajishirzi, and Ali Farhadi.
\newblock A diagram is worth a dozen images.
\newblock In \emph{Proceedings of the 14th European Conference on Computer Vision (ECCV)}, pages 235--251, 2016.

\bibitem[Lee et~al.(2024)Lee, Park, Won~Kim, and Man~Ro]{lee2024moaimixtureintelligencelarge}
Byung-Kwan Lee, Beomchan Park, Chae Won~Kim, and Yong Man~Ro.
\newblock {MoAI}: Mixture of all intelligence for large language and vision models.
\newblock In \emph{Proceedings of the 18th European Conference on Computer Vision (ECCV)}, pages 273--302, 2024.

\bibitem[Lee et~al.(2023)Lee, Joshi, Turc, Hu, Liu, Eisenschlos, Khandelwal, Shaw, Chang, and Toutanova]{lee2023pix2struct}
Kenton Lee, Mandar Joshi, Iulia~Raluca Turc, Hexiang Hu, Fangyu Liu, Julian~Martin Eisenschlos, Urvashi Khandelwal, Peter Shaw, Ming-Wei Chang, and Kristina Toutanova.
\newblock {Pix2Struct}: Screenshot parsing as pretraining for visual language understanding.
\newblock In \emph{Proceedings of the 40th International Conference on Machine Learning (ICML)}, pages 18893--18912, 2023.

\bibitem[Li et~al.(2024{\natexlab{a}})Li, Wang, Wang, Ge, Ge, and Shan]{li2023seed}
Bohao Li, Rui Wang, Guangzhi Wang, Yuying Ge, Yixiao Ge, and Ying Shan.
\newblock {SEED-Bench}: Benchmarking multimodal large language models.
\newblock In \emph{2024 IEEE/CVF Conference on Computer Vision and Pattern Recognition (CVPR)}, pages 13299--13308, 2024{\natexlab{a}}.

\bibitem[Li et~al.(2025{\natexlab{a}})Li, Zhang, Liao, Peng, Ding, and Jin]{li2025redundancylensrevealingexploitingvisual}
Hongliang Li, Jiaxin Zhang, Wenhui Liao, Dezhi Peng, Kai Ding, and Lianwen Jin.
\newblock {RedundancyLens}: Revealing and exploiting visual token processing redundancy for efficient decoder-only {MLLMs}.
\newblock \emph{arXiv preprint arXiv:2501.19036}, 2025{\natexlab{a}}.

\bibitem[Li et~al.(2025{\natexlab{b}})Li, Yuan, Liu, Tang, Wang, Qin, Zhu, and Zhang]{li2024tokenpacker}
Wentong Li, Yuqian Yuan, Jian Liu, Dongqi Tang, Song Wang, Jie Qin, Jianke Zhu, and Lei Zhang.
\newblock {TokenPacker}: Efficient visual projector for multimodal {LLM}.
\newblock \emph{International Journal of Computer Vision}, pages 1--19, 2025{\natexlab{b}}.

\bibitem[Li et~al.(2023)Li, Du, Zhou, Wang, Zhao, and Wen]{li2023obj_Hallucination}
Yifan Li, Yifan Du, Kun Zhou, Jinpeng Wang, Wayne~Xin Zhao, and Ji-Rong Wen.
\newblock Evaluating object hallucination in large vision-language models.
\newblock In \emph{Proceedings of the 2023 Conference on Empirical Methods in Natural Language Processing}, pages 292--305, 2023.

\bibitem[Li et~al.(2024{\natexlab{b}})Li, Zhang, Wang, Zhong, Chen, Chu, Liu, and Jia]{li2024minigeminiminingpotentialmultimodality}
Yanwei Li, Yuechen Zhang, Chengyao Wang, Zhisheng Zhong, Yixin Chen, Ruihang Chu, Shaoteng Liu, and Jiaya Jia.
\newblock {Mini-Gemini}: Mining the potential of multi-modality vision language models.
\newblock \emph{arXiv preprint arXiv:2403.18814}, 2024{\natexlab{b}}.

\bibitem[Li et~al.(2024{\natexlab{c}})Li, Yang, Liu, Ma, Zhang, Yang, Sun, Liu, and Bai]{li2024monkeyimageresolutiontext}
Zhang Li, Biao Yang, Qiang Liu, Zhiyin Ma, Shuo Zhang, Jingxu Yang, Yabo Sun, Yuliang Liu, and Xiang Bai.
\newblock Monkey: Image resolution and text label are important things for large multi-modal models.
\newblock In \emph{2024 IEEE/CVF Conference on Computer Vision and Pattern Recognition (CVPR)}, pages 26763--26773, 2024{\natexlab{c}}.

\bibitem[Lin et~al.(2020)Lin, Ji, Wang, Zhang, Zhang, Tian, and Shao]{Lin_2020_CVPR}
Mingbao Lin, Rongrong Ji, Yan Wang, Yichen Zhang, Baochang Zhang, Yonghong Tian, and Ling Shao.
\newblock {HRank}: Filter pruning using high-rank feature map.
\newblock In \emph{2020 IEEE/CVF Conference on Computer Vision and Pattern Recognition (CVPR)}, pages 1529--1538, 2020.

\bibitem[Lin et~al.(2024)Lin, Liu, Zhang, Gao, Qiu, Xiao, Qiu, Shao, Chen, Han, Huang, Zhang, He, Qiao, and Li]{lin2023sphinx}
Ziyi Lin, Dongyang Liu, Renrui Zhang, Peng Gao, Longtian Qiu, Han Xiao, Han Qiu, Wenqi Shao, Keqin Chen, Jiaming Han, Siyuan Huang, Yichi Zhang, Xuming He, Yu Qiao, and Hongsheng Li.
\newblock {SPHINX}: A mixer of weights, visual embeddings and image scales for multi-modal large language models.
\newblock In \emph{Proceedings of the 18th European Conference on Computer Vision (ECCV)}, pages 36--55, 2024.

\bibitem[Lin et~al.(2025)Lin, Lin, Lin, and Ji]{lin2024boosting}
Zhihang Lin, Mingbao Lin, Luxi Lin, and Rongrong Ji.
\newblock Boosting multimodal large language models with visual tokens withdrawal for rapid inference.
\newblock In \emph{Proceedings of the 39th AAAI Conference on Artificial Intelligence}, pages 5334--5342, 2025.

\bibitem[Liu et~al.(2023{\natexlab{a}})Liu, Li, Wu, and Lee]{liu2023visual}
Haotian Liu, Chunyuan Li, Qingyang Wu, and Yong~Jae Lee.
\newblock Visual instruction tuning.
\newblock In \emph{Advances in Neural Information Processing Systems 36}, pages 34892--34916, 2023{\natexlab{a}}.

\bibitem[Liu et~al.(2024{\natexlab{a}})Liu, Li, Li, and Lee]{liu2023improvedllava}
Haotian Liu, Chunyuan Li, Yuheng Li, and Yong~Jae Lee.
\newblock Improved baselines with visual instruction tuning.
\newblock In \emph{2024 IEEE/CVF Conference on Computer Vision and Pattern Recognition (CVPR)}, pages 26296--26306, 2024{\natexlab{a}}.

\bibitem[Liu et~al.(2024{\natexlab{b}})Liu, Li, Li, Li, Zhang, Shen, and Lee]{liu2024llavanext}
Haotian Liu, Chunyuan Li, Yuheng Li, Bo Li, Yuanhan Zhang, Sheng Shen, and Yong~Jae Lee.
\newblock {LLaVA-NeXT}: Improved reasoning, {OCR}, and world knowledge.
\newblock \url{https://llava-vl.github.io/blog/2024-01-30-llava-next/}, 2024{\natexlab{b}}.

\bibitem[Liu et~al.(2024{\natexlab{c}})Liu, Fan, Johns, Yu, Xiao, and Anandkumar]{liu2024prismervisionlanguagemodelmultitask}
Shikun Liu, Linxi Fan, Edward Johns, Zhiding Yu, Chaowei Xiao, and Anima Anandkumar.
\newblock Prismer: A vision-language model with multi-task experts.
\newblock \emph{Transactions on Machine Learning Resear}, 2024{\natexlab{c}}.

\bibitem[Liu et~al.(2024{\natexlab{d}})Liu, Shi, Hong, Hu, Yin, and Zhang]{liu2024multi}
Ting Liu, Liangtao Shi, Richang Hong, Yue Hu, Quanjun Yin, and Linfeng Zhang.
\newblock Multi-stage vision token dropping: Towards efficient multimodal large language model.
\newblock \emph{arXiv preprint arXiv:2411.10803}, 2024{\natexlab{d}}.

\bibitem[Liu et~al.(2023{\natexlab{b}})Liu, Li, Yang, Li, Yin, Liu, Jin, and Bai]{liu2023hidden}
Yuliang Liu, Zhang Li, Biao Yang, Chunyuan Li, Xucheng Yin, Cheng-lin Liu, Lianwen Jin, and Xiang Bai.
\newblock On the hidden mystery of {OCR} in large multimodal models.
\newblock \emph{arXiv preprint arXiv:2305.07895}, 2023{\natexlab{b}}.

\bibitem[Liu et~al.(2024{\natexlab{e}})Liu, Duan, Zhang, Li, Zhang, Zhao, Yuan, Wang, He, Liu, Chen, and Lin]{liu2024mmbench}
Yuan Liu, Haodong Duan, Yuanhan Zhang, Bo Li, Songyang Zhang, Wangbo Zhao, Yike Yuan, Jiaqi Wang, Conghui He, Ziwei Liu, Kai Chen, and Dahua Lin.
\newblock {MMBench}: Is your multi-modal model an all-around player?
\newblock In \emph{Proceedings of the 18th European Conference on Computer Vision (ECCV)}, pages 216--233, 2024{\natexlab{e}}.

\bibitem[Liu et~al.(2022)Liu, Mao, Wu, Feichtenhofer, Darrell, and Xie]{liu2022convnet}
Zhuang Liu, Hanzi Mao, Chao-Yuan Wu, Christoph Feichtenhofer, Trevor Darrell, and Saining Xie.
\newblock A {ConvNet} for the 2020s.
\newblock In \emph{2022 IEEE/CVF Conference on Computer Vision and Pattern Recognition (CVPR)}, pages 11976--11986, 2022.

\bibitem[Lu et~al.(2024)Lu, Liu, Zhang, Wang, Dong, Liu, Sun, Ren, Li, Yang, Sun, Deng, Xu, Xie, and Ruan]{lu2024deepseekvlrealworldvisionlanguageunderstanding}
Haoyu Lu, Wen Liu, Bo Zhang, Bingxuan Wang, Kai Dong, Bo Liu, Jingxiang Sun, Tongzheng Ren, Zhuoshu Li, Hao Yang, Yaofeng Sun, Chengqi Deng, Hanwei Xu, Zhenda Xie, and Chong Ruan.
\newblock {DeepSeek-VL}: towards real-world vision-language understanding.
\newblock \emph{arXiv preprint arXiv:2403.05525}, 2024.

\bibitem[Lu et~al.(2022)Lu, Mishra, Xia, Qiu, Chang, Zhu, Tafjord, Clark, and Kalyan]{lu2022learn}
Pan Lu, Swaroop Mishra, Tanglin Xia, Liang Qiu, Kai-Wei Chang, Song-Chun Zhu, Oyvind Tafjord, Peter Clark, and Ashwin Kalyan.
\newblock Learn to explain: Multimodal reasoning via thought chains for science question answering.
\newblock In \emph{Advances in Neural Information Processing Systems 35}, pages 2507--2521, 2022.

\bibitem[Luo et~al.(2025)Luo, Zhou, Zhang, Zheng, Sun, and Ji]{luo2024feast}
Gen Luo, Yiyi Zhou, Yuxin Zhang, Xiawu Zheng, Xiaoshuai Sun, and Rongrong Ji.
\newblock Feast your eyes: Mixture-of-resolution adaptation for multimodal large language models.
\newblock In \emph{The Thirteenth International Conference on Learning Representations}, 2025.

\bibitem[Marino et~al.(2019)Marino, Rastegari, Farhadi, and Mottaghi]{marino2019ok}
Kenneth Marino, Mohammad Rastegari, Ali Farhadi, and Roozbeh Mottaghi.
\newblock {OK-VQA}: A visual question answering benchmark requiring external knowledge.
\newblock In \emph{2019 IEEE/CVF Conference on Computer Vision and Pattern Recognition (CVPR)}, pages 3190--3199, 2019.

\bibitem[Masry et~al.(2022)Masry, Long, Tan, Joty, and Hoque]{masry2022chartqa}
Ahmed Masry, Do~Xuan Long, Jia~Qing Tan, Shafiq Joty, and Enamul Hoque.
\newblock {ChartQA}: A benchmark for question answering about charts with visual and logical reasoning.
\newblock In \emph{Findings of the Association for Computational Linguistics: ACL 2022}, pages 2263--2279, 2022.

\bibitem[Mathew et~al.(2021)Mathew, Karatzas, and Jawahar]{mathew2021docvqa}
Minesh Mathew, Dimosthenis Karatzas, and CV Jawahar.
\newblock {DocVQA}: A dataset for {VQA} on document images.
\newblock In \emph{2021 IEEE/CVF Winter Conference on Applications of Computer Vision (WACV)}, pages 2200--2209, 2021.

\bibitem[McKinzie et~al.(2024)McKinzie, Gan, Fauconnier, Dodge, Zhang, Dufter, Shah, Du, Peng, Weers, Belyi, Zhang, Singh, Kang, Jain, Hè, Schwarzer, Gunter, Kong, Zhang, Wang, Wang, Du, Lei, Wiseman, Yin, Lee, Wang, Pang, Grasch, Toshev, and Yang]{mckinzie2024mm1}
Brandon McKinzie, Zhe Gan, Jean-Philippe Fauconnier, Sam Dodge, Bowen Zhang, Philipp Dufter, Dhruti Shah, Xianzhi Du, Futang Peng, Floris Weers, Anton Belyi, Haotian Zhang, Karanjeet Singh, Doug Kang, Ankur Jain, Hongyu Hè, Max Schwarzer, Tom Gunter, Xiang Kong, Aonan Zhang, Jianyu Wang, Chong Wang, Nan Du, Tao Lei, Sam Wiseman, Guoli Yin, Mark Lee, Zirui Wang, Ruoming Pang, Peter Grasch, Alexander Toshev, and Yinfei Yang.
\newblock {MM1}: methods, analysis and insights from multimodal {LLM} pre-training.
\newblock In \emph{Proceedings of the 18th European Conference on Computer Vision (ECCV)}, pages 304--323, 2024.

\bibitem[Meng et~al.(2024)Meng, Yang, Tian, Dai, Wu, Gao, and Jiang]{meng2024deepstack}
Lingchen Meng, Jianwei Yang, Rui Tian, Xiyang Dai, Zuxuan Wu, Jianfeng Gao, and Yu-Gang Jiang.
\newblock {DeepStack}: Deeply stacking visual tokens is surprisingly simple and effective for {LMMs}.
\newblock In \emph{Advances in Neural Information Processing Systems 37}, pages 23464--23487, 2024.

\bibitem[Radford et~al.(2021)Radford, Kim, Hallacy, Ramesh, Goh, Agarwal, Sastry, Askell, Mishkin, Clark, Krueger, and Sutskever]{radford2021learning}
Alec Radford, Jong~Wook Kim, Chris Hallacy, Aditya Ramesh, Gabriel Goh, Sandhini Agarwal, Girish Sastry, Amanda Askell, Pamela Mishkin, Jack Clark, Gretchen Krueger, and Ilya Sutskever.
\newblock Learning transferable visual models from natural language supervision.
\newblock In \emph{Proceedings of the 38th International Conference on Machine Learning (ICML)}, pages 8748--8763, 2021.

\bibitem[Shang et~al.(2024)Shang, Cai, Xu, Lee, and Yan]{shang2024llava}
Yuzhang Shang, Mu Cai, Bingxin Xu, Yong~Jae Lee, and Yan Yan.
\newblock {LLaVA-PruMerge}: Adaptive token reduction for efficient large multimodal models.
\newblock \emph{arXiv preprint arXiv:2403.15388}, 2024.

\bibitem[Shi et~al.(2025)Shi, Liu, Wang, Liao, Radhakrishnan, Zhao, Huang, Yin, Sapra, Yacoob, Shi, Catanzaro, Tao, Kautz, Yu, and Liu]{shi2024eagleexploringdesignspace}
Min Shi, Fuxiao Liu, Shihao Wang, Shijia Liao, Subhashree Radhakrishnan, Yilin Zhao, De-An Huang, Hongxu Yin, Karan Sapra, Yaser Yacoob, Humphrey Shi, Bryan Catanzaro, Andrew Tao, Jan Kautz, Zhiding Yu, and Guilin Liu.
\newblock Eagle: Exploring the design space for multimodal {LLMs} with mixture of encoders.
\newblock In \emph{The Thirteenth International Conference on Learning Representations}, 2025.

\bibitem[Singh et~al.(2019)Singh, Natarajan, Shah, Jiang, Chen, Batra, Parikh, and Rohrbach]{singh2019towards}
Amanpreet Singh, Vivek Natarajan, Meet Shah, Yu Jiang, Xinlei Chen, Dhruv Batra, Devi Parikh, and Marcus Rohrbach.
\newblock Towards {VQA} models that can read.
\newblock In \emph{2019 IEEE/CVF Conference on Computer Vision and Pattern Recognition (CVPR)}, pages 8317--8326, 2019.

\bibitem[Sun et~al.(2023)Sun, Fang, Wu, Wang, and Cao]{sun2023eva}
Quan Sun, Yuxin Fang, Ledell Wu, Xinlong Wang, and Yue Cao.
\newblock {EVA-CLIP}: Improved training techniques for {CLIP} at scale.
\newblock \emph{arXiv preprint arXiv:2303.15389}, 2023.

\bibitem[Tong et~al.(2024{\natexlab{a}})Tong, Brown, Wu, Woo, Middepogu, Akula, Yang, Yang, Iyer, Pan, Wang, Fergus, LeCun, and Xie]{tong2024cambrian1fullyopenvisioncentric}
Shengbang Tong, Ellis Brown, Penghao Wu, Sanghyun Woo, Manoj Middepogu, Sai~Charitha Akula, Jihan Yang, Shusheng Yang, Adithya Iyer, Xichen Pan, Ziteng Wang, Rob Fergus, Yann LeCun, and Saining Xie.
\newblock Cambrian-1: A fully open, vision-centric exploration of multimodal {LLMs}.
\newblock In \emph{Advances in Neural Information Processing Systems 37}, pages 87310--87356, 2024{\natexlab{a}}.

\bibitem[Tong et~al.(2024{\natexlab{b}})Tong, Liu, Zhai, Ma, LeCun, and Xie]{tong2024eyeswideshutexploring}
Shengbang Tong, Zhuang Liu, Yuexiang Zhai, Yi Ma, Yann LeCun, and Saining Xie.
\newblock Eyes wide shut? exploring the visual shortcomings of multimodal {LLMs}.
\newblock In \emph{2024 IEEE/CVF Conference on Computer Vision and Pattern Recognition (CVPR)}, pages 9568--9578, 2024{\natexlab{b}}.

\bibitem[Touvron et~al.(2023)Touvron, Lavril, Izacard, Martinet, Lachaux, Lacroix, Rozière, Goyal, Hambro, Azhar, Rodriguez, Joulin, Grave, and Lample]{touvron2023llama}
Hugo Touvron, Thibaut Lavril, Gautier Izacard, Xavier Martinet, Marie-Anne Lachaux, Timothée Lacroix, Baptiste Rozière, Naman Goyal, Eric Hambro, Faisal Azhar, Aurelien Rodriguez, Armand Joulin, Edouard Grave, and Guillaume Lample.
\newblock {LLaMA}: Open and efficient foundation language models.
\newblock \emph{arXiv preprint arXiv:2302.13971}, 2023.

\bibitem[Wang et~al.(2024)Wang, Sun, Chen, Lin, Han, and Ding]{wang2024cls}
Ao Wang, Fengyuan Sun, Hui Chen, Zijia Lin, Jungong Han, and Guiguang Ding.
\newblock {[CLS]} token tells everything needed for training-free efficient {MLLMs}.
\newblock \emph{arXiv preprint arXiv:2412.05819}, 2024.

\bibitem[Wang et~al.(2025)Wang, Yu, Spadaro, Ju, Qu{\'e}tu, and Tartaglione]{wang2025folder}
Haicheng Wang, Zhemeng Yu, Gabriele Spadaro, Chen Ju, Victor Qu{\'e}tu, and Enzo Tartaglione.
\newblock {FOLDER}: Accelerating multi-modal large language models with enhanced performance.
\newblock \emph{arXiv preprint arXiv:2501.02430}, 2025.

\bibitem[Wei et~al.(2024)Wei, Kong, Chen, Zhao, Ge, Yang, Sun, Han, and Zhang]{wei2024vary}
Haoran Wei, Lingyu Kong, Jinyue Chen, Liang Zhao, Zheng Ge, Jinrong Yang, Jianjian Sun, Chunrui Han, and Xiangyu Zhang.
\newblock Vary: Scaling up the vision vocabulary for large vision-language model.
\newblock In \emph{Proceedings of the 18th European Conference on Computer Vision (ECCV)}, pages 408--424, 2024.

\bibitem[Xing et~al.(2024)Xing, Huang, Dong, Lu, Zhang, Zang, Cao, He, Wang, Wu, and Lin]{xing2024pyramiddrop}
Long Xing, Qidong Huang, Xiaoyi Dong, Jiajie Lu, Pan Zhang, Yuhang Zang, Yuhang Cao, Conghui He, Jiaqi Wang, Feng Wu, and Dahua Lin.
\newblock {PyramidDrop}: Accelerating your large vision-language models via pyramid visual redundancy reduction.
\newblock \emph{arXiv preprint arXiv:2410.17247}, 2024.

\bibitem[Yang et~al.(2025)Yang, Li, Cao, and Xu]{yang2025mitigating}
Tianyun Yang, Ziniu Li, Juan Cao, and Chang Xu.
\newblock Mitigating hallucination in large vision-language models via modular attribution and intervention.
\newblock In \emph{The Thirteenth International Conference on Learning Representations}, 2025.

\bibitem[Yao et~al.(2024)Yao, Li, Ren, Wang, Liu, Sun, and Hou]{yao2024deco}
Linli Yao, Lei Li, Shuhuai Ren, Lean Wang, Yuanxin Liu, Xu Sun, and Lu Hou.
\newblock {DeCo}: Decoupling token compression from semantic abstraction in multimodal large language models.
\newblock \emph{arXiv preprint arXiv:2405.20985}, 2024.

\bibitem[Ye et~al.(2024)Ye, Xu, Ye, Yan, Hu, Liu, Qian, Zhang, and Huang]{ye2024mplug}
Qinghao Ye, Haiyang Xu, Jiabo Ye, Ming Yan, Anwen Hu, Haowei Liu, Qi Qian, Ji Zhang, and Fei Huang.
\newblock {mPLUG-Owl2}: Revolutionizing multi-modal large language model with modality collaboration.
\newblock In \emph{2024 IEEE/CVF Conference on Computer Vision and Pattern Recognition (CVPR)}, pages 13040--13051, 2024.

\bibitem[Ye et~al.(2025{\natexlab{a}})Ye, Wu, Lin, and Zhou]{ye2024fit}
Weihao Ye, Qiong Wu, Wenhao Lin, and Yiyi Zhou.
\newblock Fit and prune: Fast and training-free visual token pruning for multi-modal large language models.
\newblock In \emph{Proceedings of the 39th AAAI Conference on Artificial Intelligence}, pages 22128--22136, 2025{\natexlab{a}}.

\bibitem[Ye et~al.(2025{\natexlab{b}})Ye, Gan, Ge, Zhang, and Tang]{ye2024atp}
Xubing Ye, Yukang Gan, Yixiao Ge, Xiao-Ping Zhang, and Yansong Tang.
\newblock {ATP-LLaVA}: Adaptive token pruning for large vision language models.
\newblock In \emph{2025 IEEE/CVF Conference on Computer Vision and Pattern Recognition (CVPR)}, pages 24972--24982, 2025{\natexlab{b}}.

\bibitem[Ye et~al.(2025{\natexlab{c}})Ye, Gan, Huang, Ge, Shan, and Tang]{ye2024voco}
Xubing Ye, Yukang Gan, Xiaoke Huang, Yixiao Ge, Ying Shan, and Yansong Tang.
\newblock {VoCo-LLaMA}: Towards vision compression with large language models.
\newblock In \emph{2025 IEEE/CVF Conference on Computer Vision and Pattern Recognition (CVPR)}, pages 29836--29846, 2025{\natexlab{c}}.

\bibitem[Yin et~al.(2025)Yin, Si, and Wang]{yinunraveling}
Hao Yin, Guangzong Si, and Zilei Wang.
\newblock Lifting the veil on visual information flow in {MLLMs}: Unlocking pathways to faster inference.
\newblock In \emph{2025 IEEE/CVF Conference on Computer Vision and Pattern Recognition (CVPR)}, pages 9382--9391, 2025.

\bibitem[Zhang et~al.(2024{\natexlab{a}})Zhang, Cheng, Lu, Zhuo, Wang, Cao, Guo, She, and Zhang]{zhang2024cls}
Qizhe Zhang, Aosong Cheng, Ming Lu, Zhiyong Zhuo, Minqi Wang, Jiajun Cao, Shaobo Guo, Qi She, and Shanghang Zhang.
\newblock {[CLS]} attention is all you need for training-free visual token pruning: Make {VLM} inference faster.
\newblock \emph{arXiv preprint arXiv:2412.01818}, 2024{\natexlab{a}}.

\bibitem[Zhang et~al.(2025{\natexlab{a}})Zhang, Fang, Yang, and Feng]{zhang2025llava}
Shaolei Zhang, Qingkai Fang, Zhe Yang, and Yang Feng.
\newblock {LLaVA-Mini}: Efficient image and video large multimodal models with one vision token.
\newblock In \emph{The Thirteenth International Conference on Learning Representations}, 2025{\natexlab{a}}.

\bibitem[Zhang et~al.(2024{\natexlab{b}})Zhang, Quan, Gu, Shen, Yuan, Yan, Cheng, Wu, and Ye]{zhang2024seeingclearlylayertwo}
Xiaofeng Zhang, Yihao Quan, Chaochen Gu, Chen Shen, Xiaosong Yuan, Shaotian Yan, Hao Cheng, Kaijie Wu, and Jieping Ye.
\newblock Seeing clearly by layer two: Enhancing attention heads to alleviate hallucination in {LVLMs}.
\newblock \emph{arXiv preprint arXiv:2411.09968}, 2024{\natexlab{b}}.

\bibitem[Zhang et~al.(2025{\natexlab{b}})Zhang, Fan, Ma, Zheng, Huang, Cheng, Gudovskiy, Okuno, Nakata, Keutzer, and Zhang]{zhang2024sparsevlm}
Yuan Zhang, Chun-Kai Fan, Junpeng Ma, Wenzhao Zheng, Tao Huang, Kuan Cheng, Denis Gudovskiy, Tomoyuki Okuno, Yohei Nakata, Kurt Keutzer, and Shanghang Zhang.
\newblock {SparseVLM}: Visual token sparsification for efficient vision-language model inference.
\newblock In \emph{Proceedings of the 42nd International Conference on Machine Learning}, 2025{\natexlab{b}}.

\bibitem[Zhang et~al.(2024{\natexlab{c}})Zhang, Pham, Zhao, Wan, Li, Zhou, Miranda, Kale, and Xu]{zhang2024treatvisualtokenstext}
Zeliang Zhang, Phu Pham, Wentian Zhao, Kun Wan, Yu-Jhe Li, Jianing Zhou, Daniel Miranda, Ajinkya Kale, and Chenliang Xu.
\newblock Treat visual tokens as text? but your {MLLM} only needs fewer efforts to see.
\newblock \emph{arXiv preprint arXiv:2410.06169}, 2024{\natexlab{c}}.

\bibitem[Zhao et~al.(2024)Zhao, Wang, Juefei-Xu, Xia, Liu, Wang, Liang, Zhang, Metaxas, and Yu]{zhao2024accelerating}
Shiyu Zhao, Zhenting Wang, Felix Juefei-Xu, Xide Xia, Miao Liu, Xiaofang Wang, Mingfu Liang, Ning Zhang, Dimitris~N Metaxas, and Licheng Yu.
\newblock Accelerating multimodel large language models by searching optimal vision token reduction.
\newblock \emph{arXiv preprint arXiv:2412.00556}, 2024.

\bibitem[Zhong et~al.(2024)Zhong, Liu, Li, and Wang]{zhong2024aim}
Yiwu Zhong, Zhuoming Liu, Yin Li, and Liwei Wang.
\newblock {AIM}: Adaptive inference of multi-modal {LLMs} via token merging and pruning.
\newblock \emph{arXiv preprint arXiv:2412.03248}, 2024.

\bibitem[Zhu et~al.(2023)Zhu, Chen, Shen, Li, and Elhoseiny]{zhu2023minigpt}
Deyao Zhu, Jun Chen, Xiaoqian Shen, Xiang Li, and Mohamed Elhoseiny.
\newblock {MiniGPT-4}: Enhancing vision-language understanding with advanced large language models.
\newblock \emph{arXiv preprint arXiv:2304.10592}, 2023.

\bibitem[Zhu et~al.(2024)Zhu, Xie, Liang, Zheng, and Guo]{zhu2024focusllava}
Yuke Zhu, Chi Xie, Shuang Liang, Bo Zheng, and Sheng Guo.
\newblock {FocusLLaVA}: A coarse-to-fine approach for efficient and effective visual token compression.
\newblock \emph{arXiv preprint arXiv:2411.14228}, 2024.

\bibitem[Zong et~al.(2024)Zong, Ma, Shen, Song, Shao, Jiang, Li, and Liu]{zong2024mova}
Zhuofan Zong, Bingqi Ma, Dazhong Shen, Guanglu Song, Hao Shao, Dongzhi Jiang, Hongsheng Li, and Yu Liu.
\newblock {MoVA}: Adapting mixture of vision experts to multimodal context.
\newblock In \emph{Advances in Neural Information Processing Systems 37}, pages 103305--103333, 2024.

\end{thebibliography}
}

\end{document}